\documentclass[]{trbunofficial}
\usepackage{graphicx}
\usepackage{xcolor}
\usepackage{soul}
\usepackage{multirow}

\usepackage[hidelinks]{hyperref}
\usepackage{subcaption}
\newcommand{\etal}{\textit{et al.}}
\newcommand{\ie}{\textit{i.e.}}
\newcommand{\eg}{\textit{e.g.}}

\AuthorHeaders{Ma, Enan, Cheng, and Chowdhury}
\title{Understanding the Risks of Asphalt Art to the Reliability of Vision-Based Perception Systems}

\author{%
  \textbf{Jin Ma\footnote{Co-first authors.}}\\
  School of Computing\\
  Clemson University, Clemson, South Carolina, 29634\\
  Email: jin7@clemson.edu\\
  \hfill\break
  \textbf{Abyad Enan\footnotemark[1]}\\
  Glenn Department of Civil Engineering\\
  Clemson University, Clemson, South Carolina, 29634\\
  Email: aenan@clemson.edu\\
  \hfill\break%
  \textbf{Long Cheng, Ph.D.}\\
  School of Computing\\
  Clemson University, Clemson, South Carolina, 29634\\
  Email: lcheng2@clemson.edu\\
  \hfill\break%
  \textbf{Mashrur Chowdhury, Ph.D.}\\
  Glenn Department of Civil Engineering\\
  Clemson University, Clemson, South Carolina, 29634\\
  Email: mac@clemson.edu
}

\WordsPerTable{250}

\TotalWords{8019}

\begin{document}
\maketitle
\section{Abstract}
Artistic crosswalks featuring asphalt arts\footnote{Throughout this paper, we use the term \textbf{asphalt art} to refer broadly to decorative or artistic patterns painted on roadway surfaces, including crosswalk murals and intersection artwork.}, introduced by different organizations in recent years, aim to enhance the visibility and safety of pedestrians. However, their visual complexity may interfere with surveillance systems that rely on vision-based object detection models. In this study, we investigate the impact of asphalt art on pedestrian detection performance of a pretrained vision-based object detection model. We construct realistic crosswalk scenarios by compositing various street art patterns into a fixed surveillance scene and evaluate the model’s performance in detecting pedestrians on asphalt-arted crosswalks under both benign and adversarial conditions. A benign case refers to pedestrian crosswalks painted with existing normal asphalt art, whereas an adversarial case involves digitally crafted or altered asphalt art perpetrated by an attacker. Our results show that while simple, color-based designs have minimal effect, complex artistic patterns, particularly those with high visual salience, can significantly degrade pedestrian detection performance. Furthermore, we demonstrate that adversarially crafted asphalt art can be exploited to deliberately obscure real pedestrians or generate non-existent pedestrian detections. These findings highlight a potential vulnerability in urban vision-based pedestrian surveillance systems, and underscore the importance of accounting for environmental visual variations when designing robust pedestrian perception models.


\hfill\break%
\noindent\textit{Keywords}: Adversarial attack, Asphalt art, Pedestrian detection, Vision-based object detection.
\newpage

\section{Introduction}
\begin{figure}[b]
    \centering
    \begin{subfigure}[b]{0.4\textwidth}
        \includegraphics[width=\linewidth]{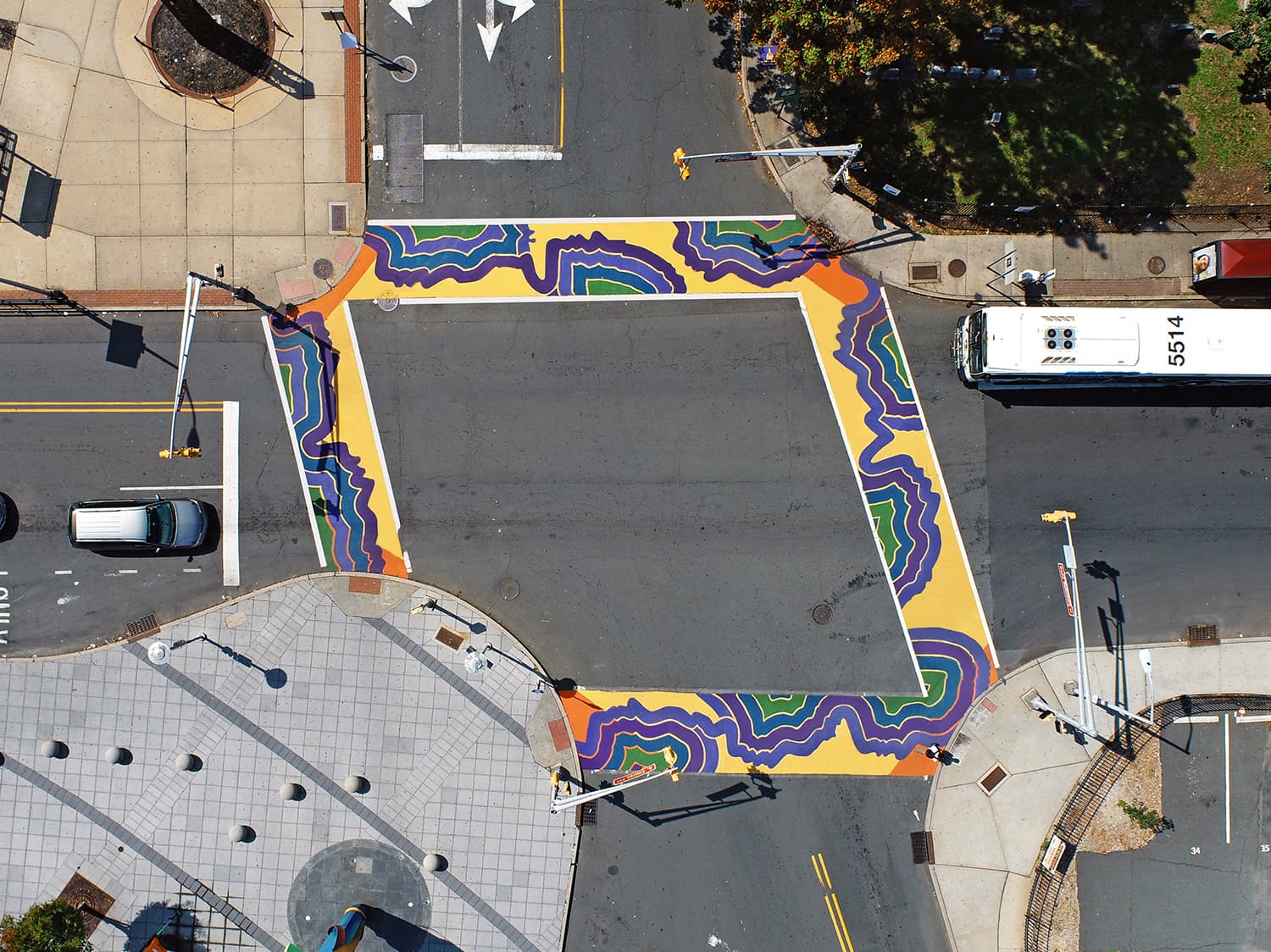}
        \caption{Asphalt art crosswalk in  New Jersey}
    \end{subfigure}
    \hfill
    \begin{subfigure}[b]{0.54\textwidth}
        \includegraphics[width=\linewidth]{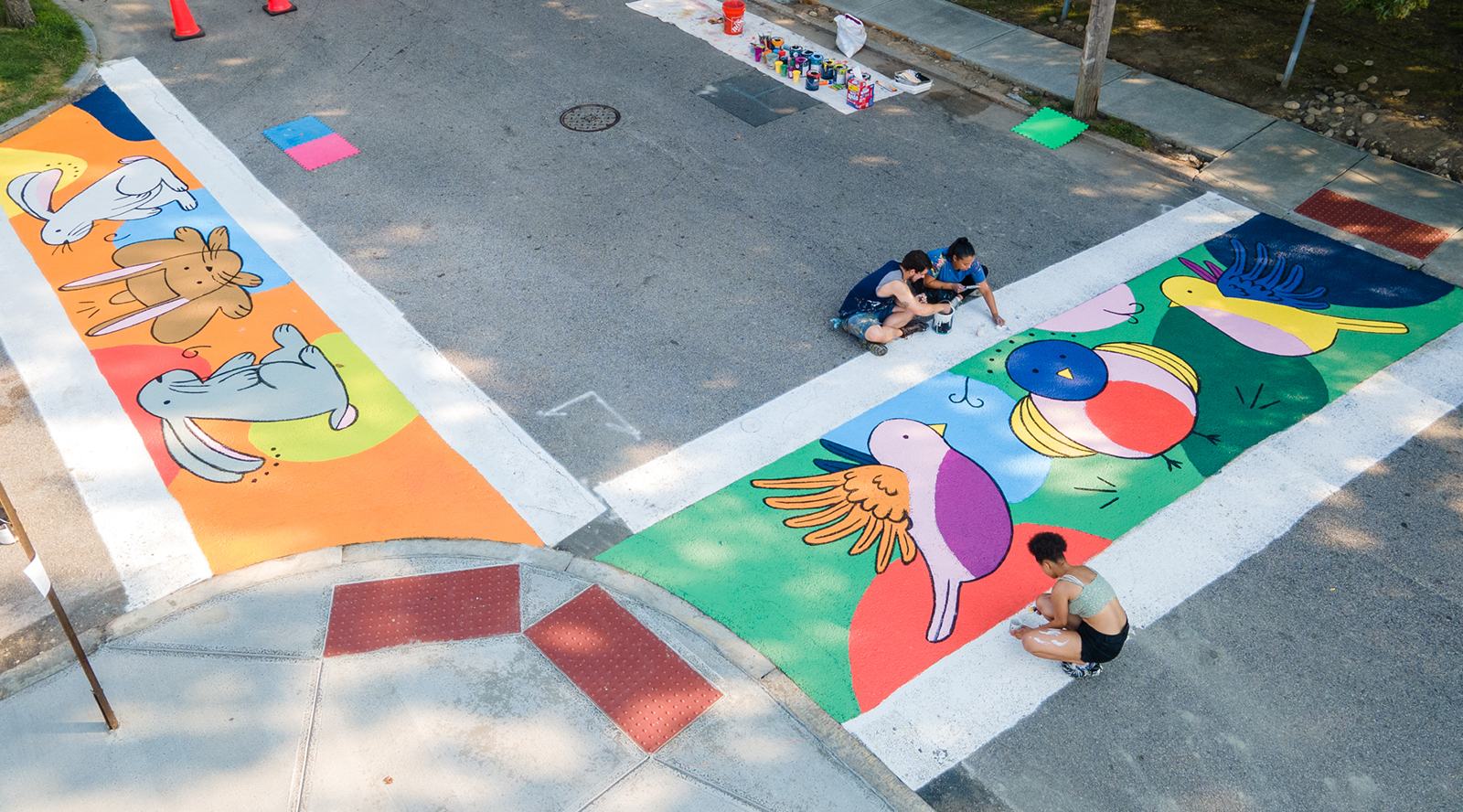}
        \caption{Asphalt art crosswalk in Rhode Island}
    \end{subfigure}
    \caption{~Examples of real-world asphalt art installations (adopted from~\cite{bloomberg}).}
    \label{fig:examples}
\end{figure}

Cities around the world have increasingly embraced asphalt art as a means to enhance road safety and street vitality, where vibrant murals are painted on crosswalks or intersections (Figure~\ref{fig:examples}). These projects aim to create highly visible, welcoming pedestrian spaces, making crosswalks more conspicuous to drivers and encouraging safer driving behaviors~\cite{asphaltart2022}. A recent study of 17 asphalt art sites recorded a 50\% drop in crashes involving pedestrians after the artwork was installed, alongside significant reductions in overall crashes and dangerous driving conflicts~\cite{asphaltart2022}. This trend suggests that creative street art can serve as a low-cost traffic-calming tool to improve pedestrian safety and urban aesthetics simultaneously. 

The growing adoption of asphalt art has prompted clarifications in traffic control regulations to ensure that safety and accessibility standards are maintained. According to the 11th Edition of the United States Manual on Uniform Traffic Control Devices (MUTCD), effective January 2024, aesthetic street treatments such as murals are permitted on roadways, provided they do not interfere with official traffic control devices, mimic regulatory pavement markings, or reduce contrast needed for crosswalk visibility and accessibility compliance~\cite{FHWA2023}. The MUTCD stipulates that non-standard colors and patterns may be applied for decorative purposes only when they do not compromise the legibility of required traffic signs and markings~\cite{FHWA2023}. The updated MUTCD explicitly allows such treatments under local agency oversight, with the expectation that jurisdictions assess potential safety impacts before installation~\cite{Bloomberg2025}. These guidelines ensure that while asphalt art can be used to enhance street vitality, it is implemented in a manner that preserves roadway clarity and minimizes confusion for all users, particularly vulnerable pedestrians.

Concurrently, advanced vision perception methods, such as You Only Look Once (YOLO) family~\cite{redmon2016you, song2023yolov5, wang2023yolov7}, have become integral components of modern artificial intelligence (AI) systems, including those used for traffic surveillance~\cite{enan2024basic}, autonomous driving~\cite{10431781}, pedestrian detection~\cite{9043590}, and other road users in real time applications~\cite{zou2023object}. Their ability to extract actionable information from complex urban environments has become foundational to intelligent transportation systems, automated safety analytics, and data-driven urban planning. However, such detectors are typically trained on large datasets of normal road scenes, so unconventional visuals, like brightly-painted crosswalk patterns, present a new scenario. It is important to ask whether these artistic roadway designs might confuse vision perception algorithms. In other words, \textit{to what extent might a benign vibrant pavement mural disrupt a vision system’s ability to accurately recognize a pedestrian crossing the street?} This question is not merely of academic interest, but also directly relevant to policy decisions in the development of smart cities, where transportation agencies must ensure that creative street treatments remain compatible with the AI systems.

Robustness to novel environmental patterns is a growing concern in the computer vision community, especially given the known vulnerability of deep learning models to adversarial perturbations~\cite{akhtar2018threat}. Prior research has shown that strategically altering the appearance of objects or surfaces can deceive even advanced detectors. For example, researchers demonstrated that a person could hide from a YOLO-based surveillance camera by holding up a small poster printed with a special pattern, \ie, the detector no longer detects the person at all~\cite{thys2019fooling}. These so-called adversarial patches exploit blind spots in vision models, and attackers could maliciously use innocuous-looking art or graffiti as camouflage. This leads to a second concern: \textit{could decorative crosswalk paintings be deliberately engineered as adversarial attacks, causing detectors to miss or misclassify pedestrians?}

In this work, we center our investigation on the above two core questions. These questions highlight an emerging security and safety concern introduced by asphalt art: while designed for community enhancement, such artwork may unintentionally disrupt or even be exploited to deceive modern vision perception systems. The failure of these systems can lead to severe consequences, particularly in safety-critical domains. For example, consider a smart intersection where infrastructure-mounted pedestrian detection models are designed to signal warnings to oncoming autonomous vehicles. If both the infrastructure’s vision system and the vehicle’s onboard perception model fail to correctly detect a pedestrian due to challenging visual conditions, such as disruptive asphalt art, the result could be a collision and serious harm to pedestrians~\cite{combs2019automated,wang2023pedestrian}. Such incidents not only endanger human life but also undermine the reliability of intelligent transportation systems~\cite{hong2020artificial}. By systematically examining how both benign and maliciously crafted asphalt art affects vision perception models, our study provides essential evidence for understanding this risk. These insights are crucial for guiding future model development, informing municipal design and approval processes for street art, and shaping policies that ensure emerging urban aesthetics remain compatible with the operational needs of AI-driven transportation systems.
\section{Background and Related Work}

\subsection{Pedestrian Detection in Vison-Based Systems}
In modern cities, vision-based perception systems increasingly rely on automatic pedestrian detection powered by deep learning. Single-stage object detectors such as YOLO~\cite{redmon2016you} and Single Shot Detector (SSD)~\cite{liu2016ssd} are especially popular for this purpose because they operate in real-time with a single network pass of an image frame. This makes it feasible to analyze multiple camera feeds simultaneously and detect people in live video streams. Such detectors have achieved high accuracy in the surveillance domain; for example, a recent YOLOv5-based model attained 96.5\% mean average precision (mAP) in pedestrian detection while processing over 50 city cameras in real time~\cite{song2023yolov5}. Integrating these intelligent detection models into surveillance not only reduces labor but also enhances security. The system can continuously monitor 24/7 and alert human operators only when a relevant target (\eg, a person in a restricted area) is detected. This approach improves efficiency and safety by automating the identification of pedestrians and potential threats in urban environments. In another study, Islam~\etal~\cite{9043590} developed a vision-based personal safety message generation method for connected vehicles, triggered when pedestrians are detected in a signalized crosswalk using a roadside pedestrian camera. Their approach utilized the YOLOv3 model for pedestrian detection and achieved a mAP exceeding 95\%. However, the robustness of such vision-based detectors is critical, as any failure to detect pedestrians, such as due to environmental factors or malicious interference, could undermine their utility in safety-critical applications.

\subsection{Attacks on Pedestrian Detection}
Although state-of-the-art detectors like YOLO are highly accurate, researchers have shown that they can be fooled by adversarial attacks in the scene. For example, Thys \etal~\cite{thys2019fooling} developed a 40×40 cm printed patch that a person can hold in front of their body to become invisible to a YOLOv2-based person detector. Wu \etal~\cite{wu2020making} presented a systematic study of wearable adversarial patterns, also referred to as an invisibility cloak. Their experiments showed that a person wearing adversarially printed clothes could evade detection by state-of-the-art models. A recent line of work focuses on making adversarial patches blend into the environment so that they appear benign to human observers. Hu \etal~\cite{hu2021naturalistic} developed a naturalistic adversarial patch generation method using generative models. By sampling from the image manifold of a Generative Adversarial Network (GAN), they crafted patches that are both effective and natural-looking (\eg, a picture of cats or dogs), making them less attention-grabbing in real scenes. 

All the above physical attacks share a common approach: the adversarial pattern is directly applied on the object. In other words, the person is modified, either wearing a special shirt or holding a patch to evade detection. By contrast, our approach is fundamentally different: we place the adversarial pattern in the background as part of the scene (\eg, as a piece of street art on the ground), without altering the person at all. This scenario simulates an attack where the environment is engineered to confuse the detector, even though the person themselves looks ordinary. To the best of our knowledge, such background-based adversarial interference with pedestrian detectors has not been explored in prior work, making this work a novel departure from existing adversarial patch attacks on surveillance systems.

\subsection{Asphalt Art Study}
Prior work on asphalt art has primarily focused on its impact on human perception, urban environments, and traffic safety. In particular, studies on urban murals and asphalt art suggest that such interventions can influence driver behavior and potentially improve safety outcomes by promoting traffic calming and increasing pedestrian awareness. For example, empirical analyses of asphalt art report reductions in crash rates and improvements in driver yielding behavior in traffic~\cite{cruz2022impact,asphaltart2022}, although certain designs may also introduce visual distraction depending on their complexity and placement~\cite{cruz2022impact}. Beyond traffic safety, asphalt art and street art have also been studied as a form of public-space intervention that reshapes how urban environments are perceived and experienced. Visconti \etal~\cite{visconti2010street} show that street art transforms anonymous spaces into meaningful places by fostering community engagement, dialogue, and a sense of collective ownership. However, these works largely examine asphalt art from a human-centric perspective and do not consider their impact on vision-based perception systems. In contrast, our work investigates asphalt art as a visual factor that can affect pedestrian detection performance in vision-based systems. This perspective highlights a previously overlooked risk, where patterns designed for aesthetic or social purposes may inadvertently interfere with automated perception, or even be exploited as adversarial designs.
\section{Research Questions and Contributions}
In this paper, we aim to investigate the potential risks that asphalt arts pose to surveillance-based pedestrian perception systems. Specifically, we seek to answer the following research questions:

\begin{enumerate}
\item Will benign asphalt art patterns on crosswalks degrade the performance of pedestrian detection models when individuals walk over them?
\item Can such crosswalk art be intentionally manipulated by a malicious actor to mislead or suppresses pedestrian detection?
\end{enumerate}

To answer those research questions, we collected a real-world surveillance perception dataset and performed both benign and malicious asphalt art simulations on this dataset. We focus on the YOLOv7 model~\cite{wang2023yolov7}, which is YOLO version 7, the seventh major iteration in the YOLO family of object detection models. We approach the problem through digital asphalt art simulation, incorporating various art patterns into street scenes observed by an overhead surveillance camera as in the collected dataset. Several case studies are considered, including real-world benign designs (\eg, colorful geometric patterns on crosswalks) as well as contrived high-contrast or noise patterns that might constitute a deliberate attack. For each scenario, we evaluate YOLOv7’s performance in detecting pedestrians, compared to a regular, unpainted crosswalk baseline. By conducting these controlled simulations, we can systematically analyze the impact of different patterns without endangering real pedestrians. Our experiments show that some artistic crosswalk designs can impair detector performance, and that deliberately crafted patterns can have even more severe effects. This study highlights critical vulnerabilities in surveillance-based perception systems and calls for more robust perception systems designed to withstand diverse environmental conditions. The key contributions of this study can be summarized as:
\begin{enumerate}
    \item Generate a surveillance-based pedestrian dataset, captured at a real-world pedestrian safety application testbed intersection, comprising over 500 manually annotated frames.
    \item Develop a photorealistic asphalt art injection pipeline, which embeds nine real-world street art patterns into crosswalk scenes to assess their effect on pedestrian detection.
    \item Perform a white-box adversarial art simulation, featuring both universal noise overlays and from-scratch pattern synthesis to systematically degrade pedestrian detection performance.
    \item Demonstrate how both benign and malicious street art interventions can compromise surveillance‐based pedestrian perception.
\end{enumerate}
\section{Research Approaches}
In this section, we outline our research methodology: first, we describe the collection and annotation of a real-world surveillance dataset; then, we detail our pipeline for compositing benign asphalt art crosswalks onto the collected dataset; finally, we show how we craft the malicious adversarial art crosswalks to attack the pedestrian detection methods.
\begin{figure}[t]
  \centering
  \begin{subfigure}[b]{0.47\textwidth}
    \centering
    \includegraphics[width=\textwidth]{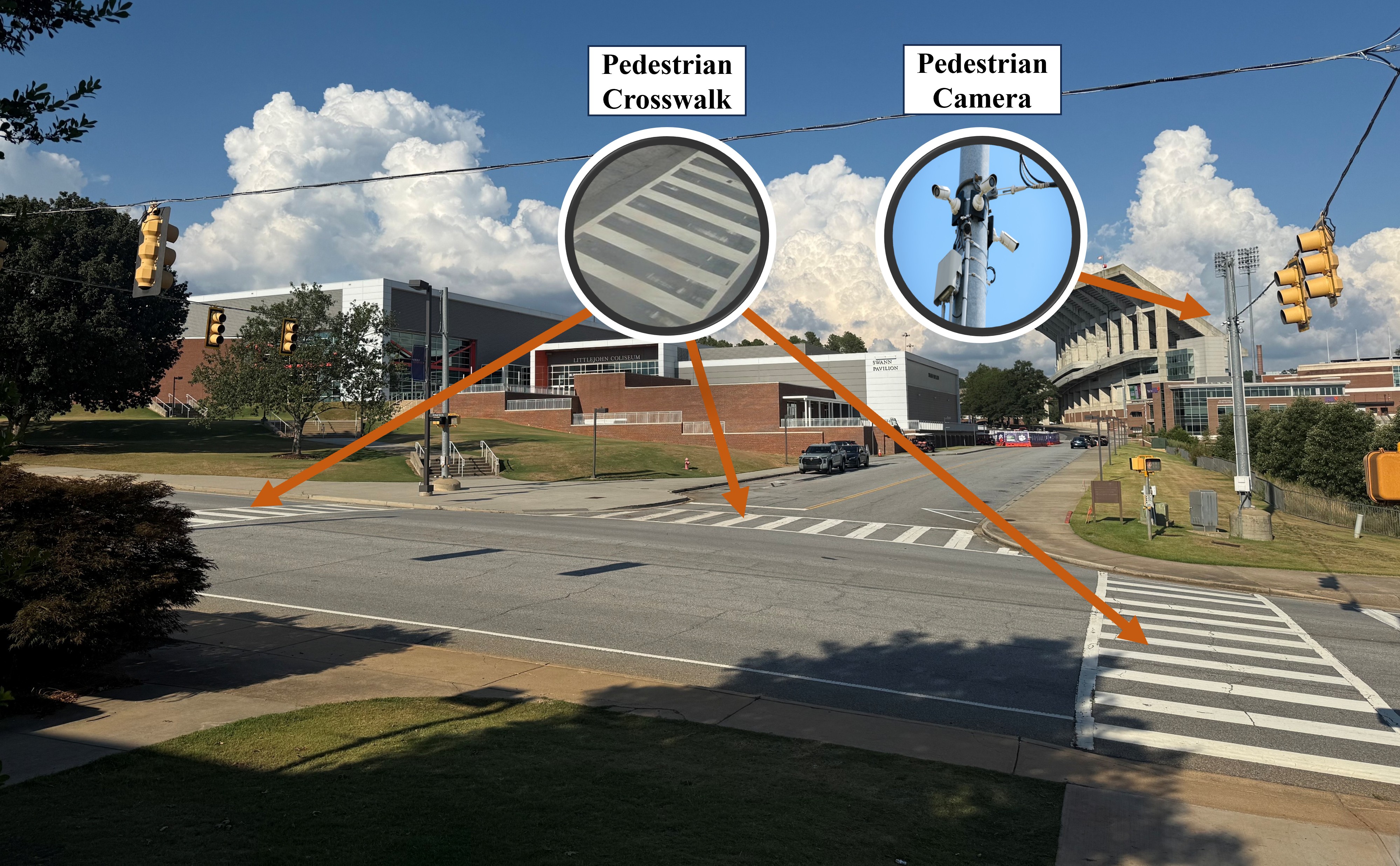}
    \caption{Testbed setting.}
    \label{testbedd}
  \end{subfigure}
  \hfill
  \begin{subfigure}[b]{0.517\textwidth}
    \centering
    \includegraphics[width=\textwidth]{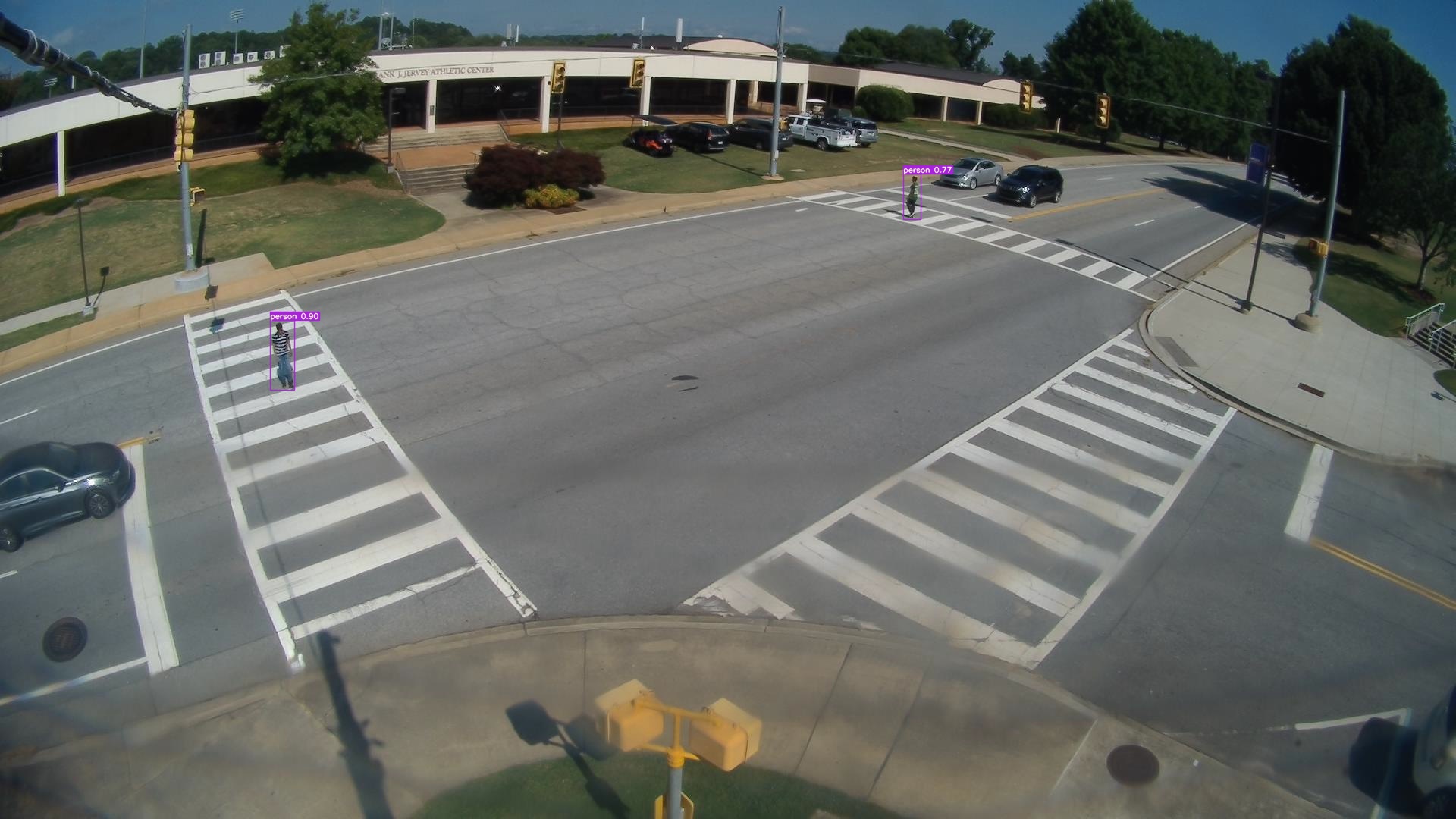}
    \caption{Example of captured image.}
    \label{camview}
  \end{subfigure}
  \caption{~Pedestrian alert warning testbed showing (a) the pedestrian crosswalk along with a pedestrian camera, and (b) an example of a captured image with the YOLOv7 pedestrian detection results.}
  \label{fig:testcam}
\end{figure}

\subsection{Data Collection and Annotation}
We collected pedestrian data at a crosswalk located at the T-intersection of Perimeter Road and the Avenue of Champions on the Clemson University campus in Clemson, South Carolina. This site is part of the South Carolina Connected Vehicle Testbed (SC-CVT)~\cite{chowdhury2018lessons}. The intersection includes three pedestrian crosswalks, as illustrated in Figure~\ref{fig:testcam}. A pedestrian camera installed at the intersection, which is part of the testbed, is connected to the Clemson University network and used to capture video data.

Although the intersection is part of the SC-CVT, it functions as a typical urban T-intersection, serving regular vehicular and pedestrian traffic. It consists of a major road intersecting a minor road and operates under a semi-actuated traffic control system, where a loop detector is installed on the minor approach and pedestrian push buttons are provided for all crossing movements. Signal phase and timing adjustments occur based on the presence of vehicles on the minor road, detected by the loop detector, and pedestrian calls, similar to the operation of conventional urban intersections. The intersection does not include any additional infrastructure that would alter natural traffic behavior. Therefore, the traffic characteristics observed at this site are representative of typical real-world urban intersection operations.

Pedestrian video footage was collected across four sessions, each lasting 15 minutes, totaling one hour of observation. The sessions were conducted during both peak and off-peak hours to capture a representative range of traffic and pedestrian conditions. During the peak-hour sessions, the average vehicular traffic volume, in each direction of the two-way roadway, was approximately 176 vehicles/hour/lane on the major road and 21 vehicles/hour/lane on the minor road, with a pedestrian volume of 66 pedestrians/hour/crosswalk. During off-peak periods, vehicular volumes averaged 159 vehicles/hour/lane on the major road and 16 vehicles/hour/lane on the minor road, with a pedestrian volume of 56 pedestrians/hour/crosswalk. All video recordings were obtained in clear daylight conditions. The video footage depicts a range of real-world pedestrian scenarios, including single pedestrians, pedestrians in groups, and individuals waiting at the crosswalk. From the video feed, image frames were extracted at a rate of two frames per second. After discarding frames without pedestrians, we retained a total of 514 images. These images were manually annotated with bounding boxes and saved accordingly. The dataset was split into training, validation, and testing subsets in a 70\%-15\%-15\% ratio, resulting in 360 images for training, 77 for validation, and 77 for testing. We trained a YOLOv7 model on this pedestrian dataset, using a pretrained backbone while retaining only the "person" class, making it a single-class object detector. This focused approach improves pedestrian detection accuracy, which aligns with the practical goal of pedestrian surveillance systems, \ie, detecting pedestrians exclusively. 


\subsection{Benign Asphalt Art Creation}
\begin{figure}[t]
    \centering
    \includegraphics[width=0.95\linewidth]{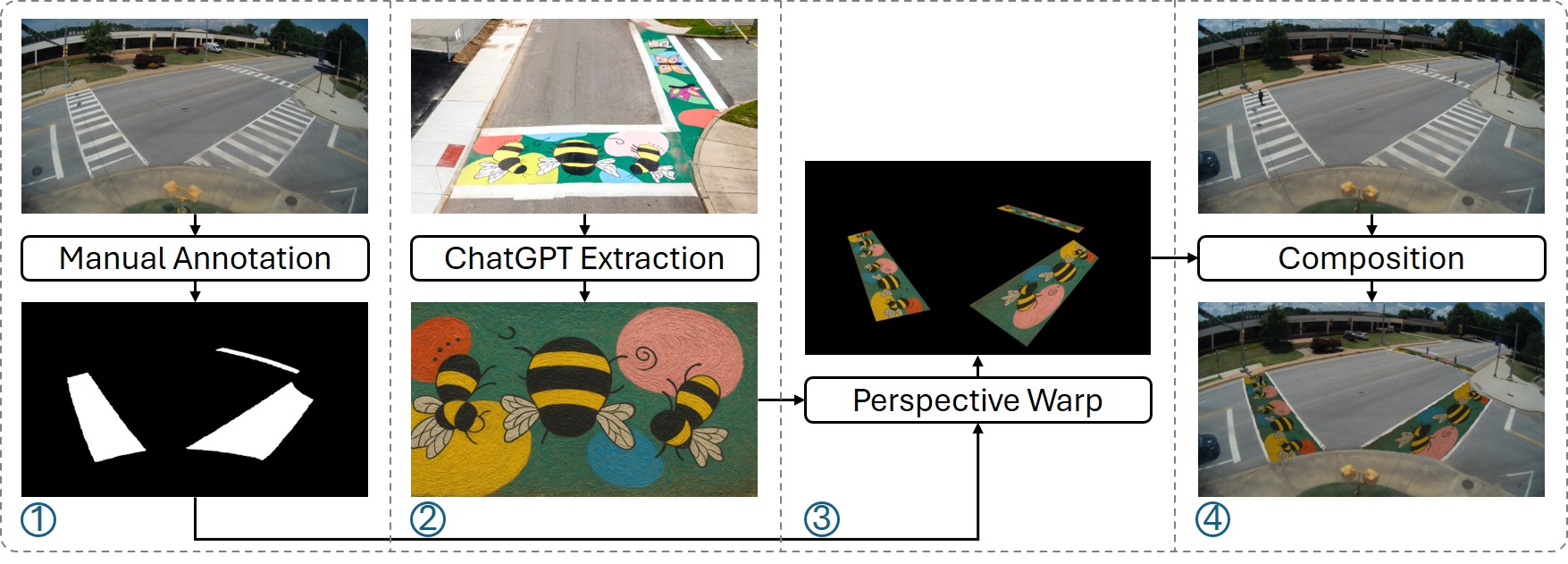}
    \caption{~The image creation process for benign asphalt art evaluation, including \textcircled{1} crosswalk mask annotation, \textcircled{2} asphalt art selection, \textcircled{3} perspective warp transformation, and \textcircled{4} applying to testing images.}
    \label{fig:pipeline}
\end{figure}

To simulate crosswalk scenarios featuring asphalt art using our collected dataset, we employ a four-step pipeline. The complete process is illustrated in Figure~\ref{fig:pipeline}.

\textbf{1. Crosswalk mask annotation.} The first step in our pipeline involves identifying the crosswalk regions where asphalt art will be applied. Since the surveillance camera in our dataset remains stationary throughout, the crosswalk positions are fixed across all frames. Consequently, we only need to annotate a single reference image without any foreground objects (e.g., pedestrians or vehicles). Using the open-source annotation tool Labelme~\cite{labelme}, we manually labeled the three crosswalks in the selected intersection scene, generating pixel-wise masks that delineate the exact regions for potential art placement. 

\begin{figure}[t]
    \centering
    \begin{subfigure}[c]{0.3\textwidth}
        \includegraphics[width=\linewidth]{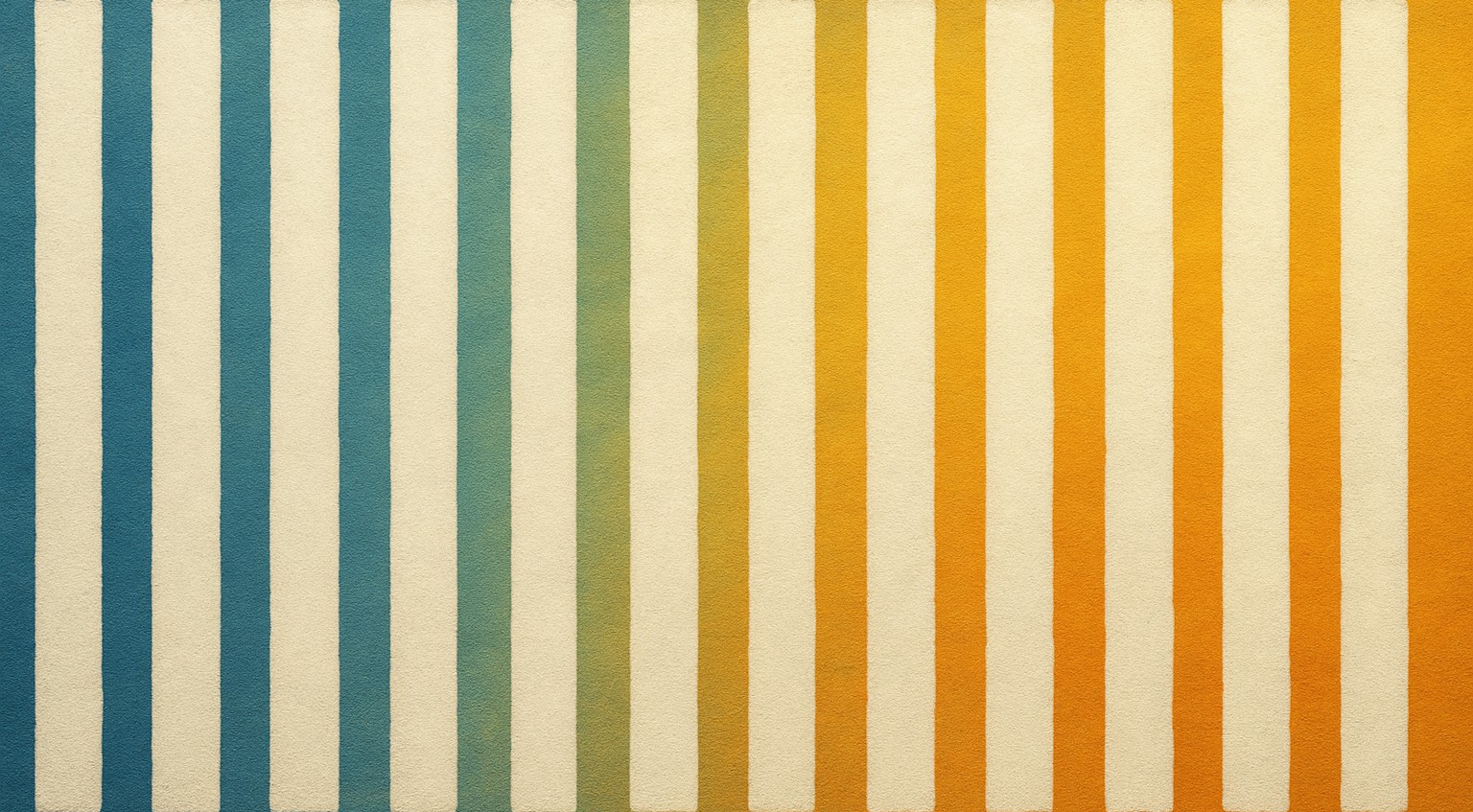}
        \caption*{(i) Color 1}
    \end{subfigure}
    \quad
    \begin{subfigure}[c]{0.3\textwidth}
        \includegraphics[width=\linewidth]{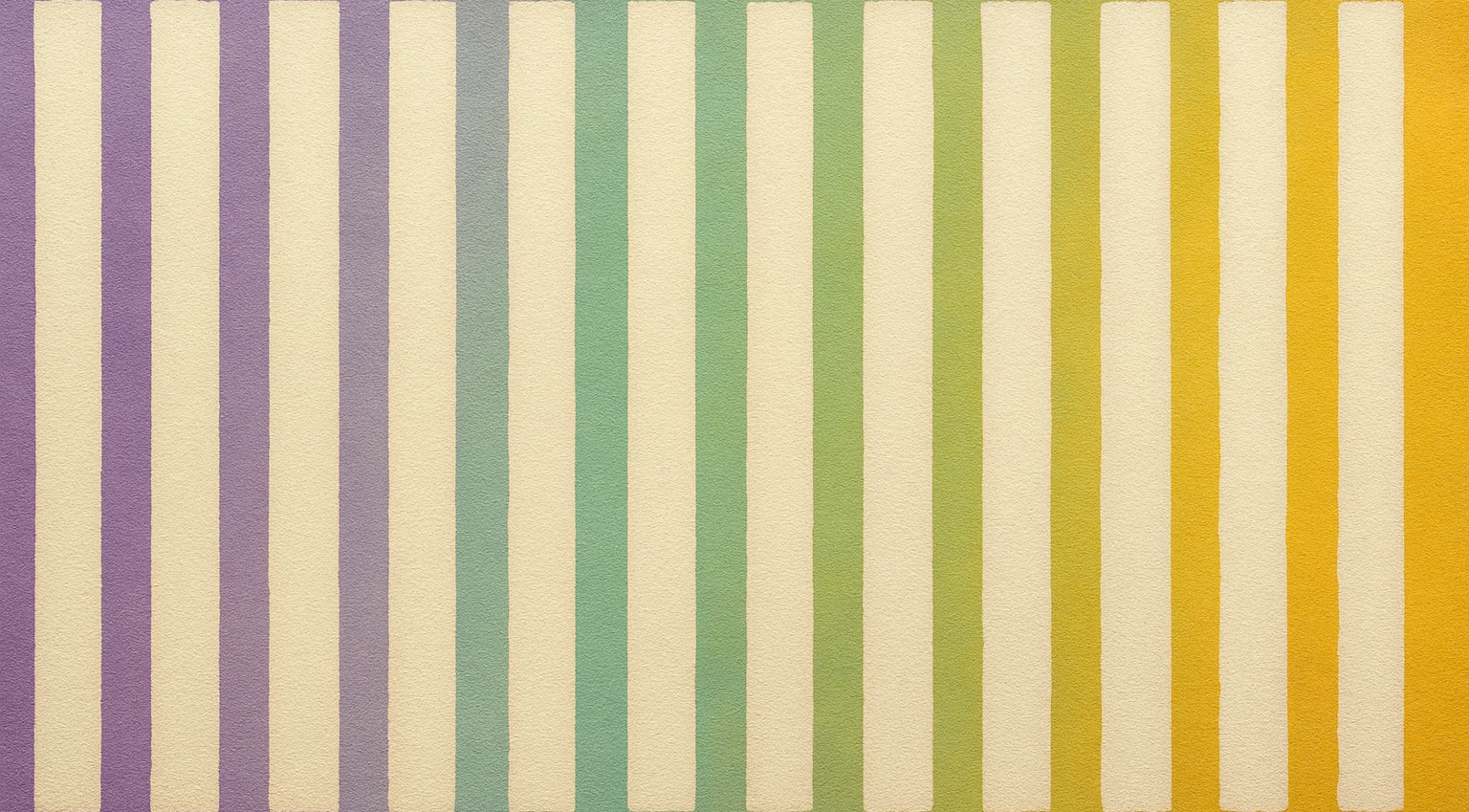}
        \caption*{(i) Color 2}
    \end{subfigure}
    \quad
    \begin{subfigure}[c]{0.3\textwidth}
        \includegraphics[width=\linewidth]{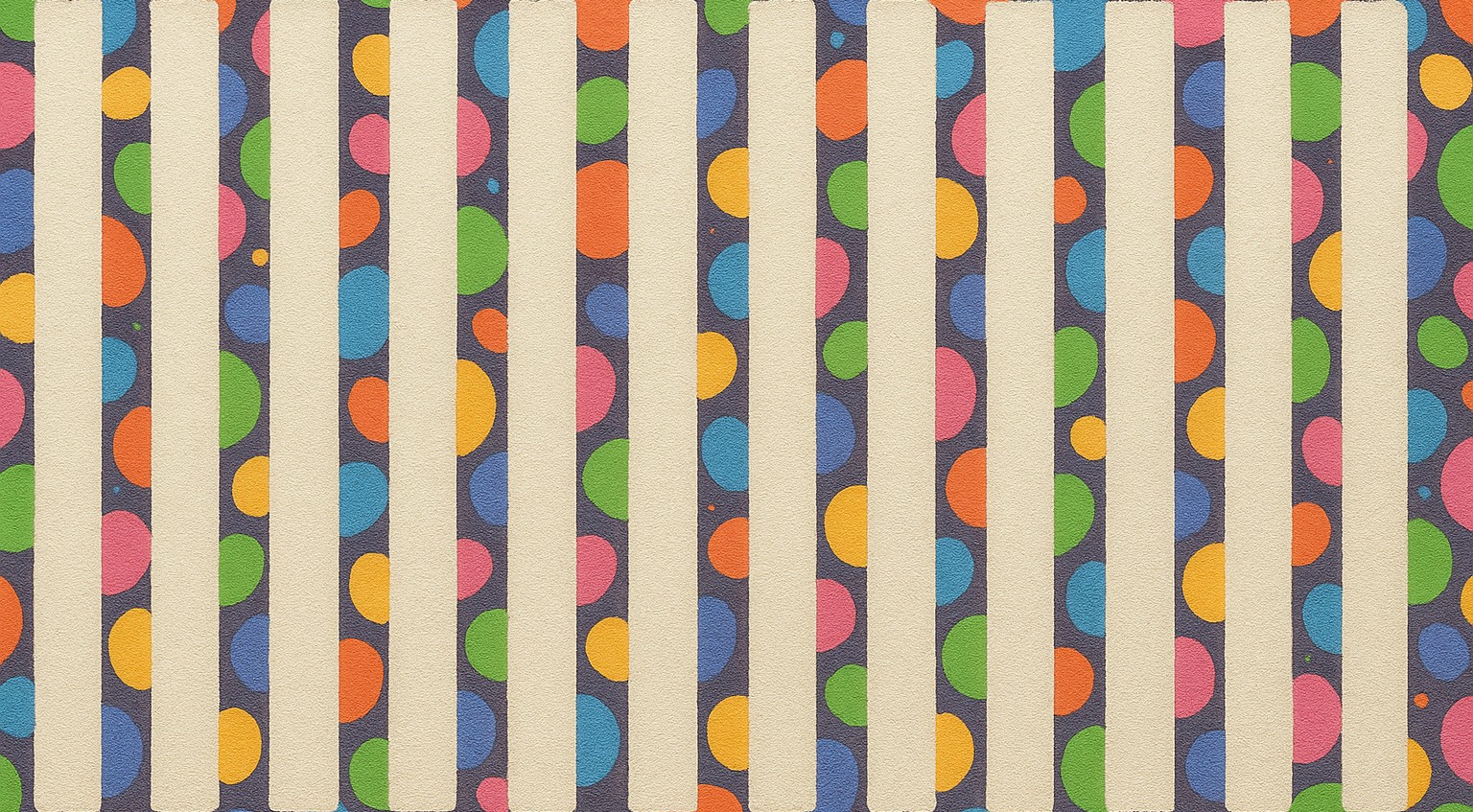}
        \caption*{(ii) Pattern 1}
    \end{subfigure}

    \vspace{8pt}

    \begin{subfigure}[c]{0.3\textwidth}
        \includegraphics[width=\linewidth]{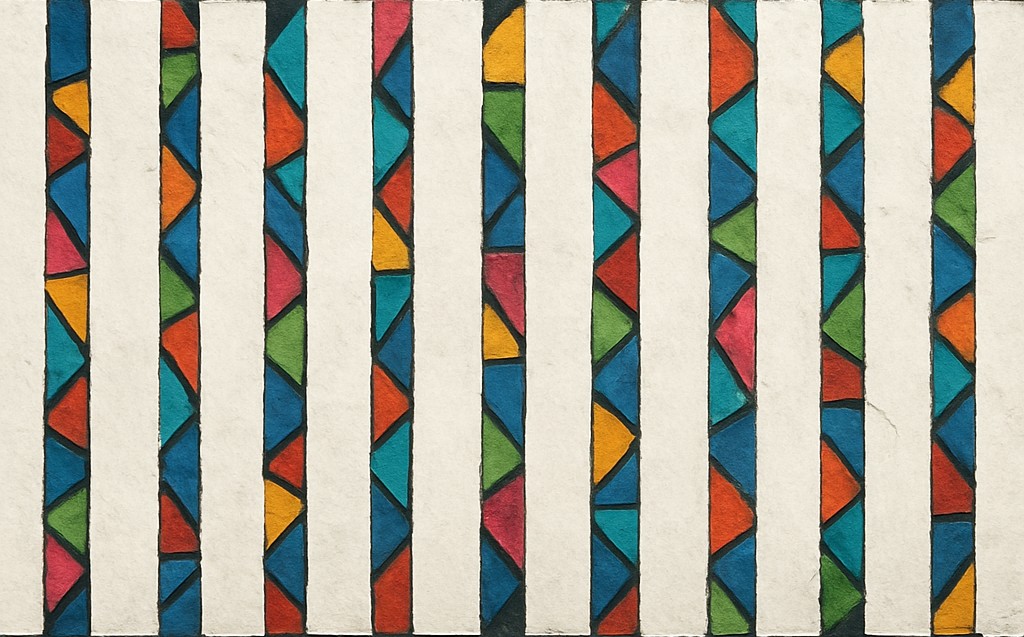}
        \caption*{(ii) Pattern 2}
    \end{subfigure}
    \quad
    \begin{subfigure}[c]{0.3\textwidth}
        \includegraphics[width=\linewidth]{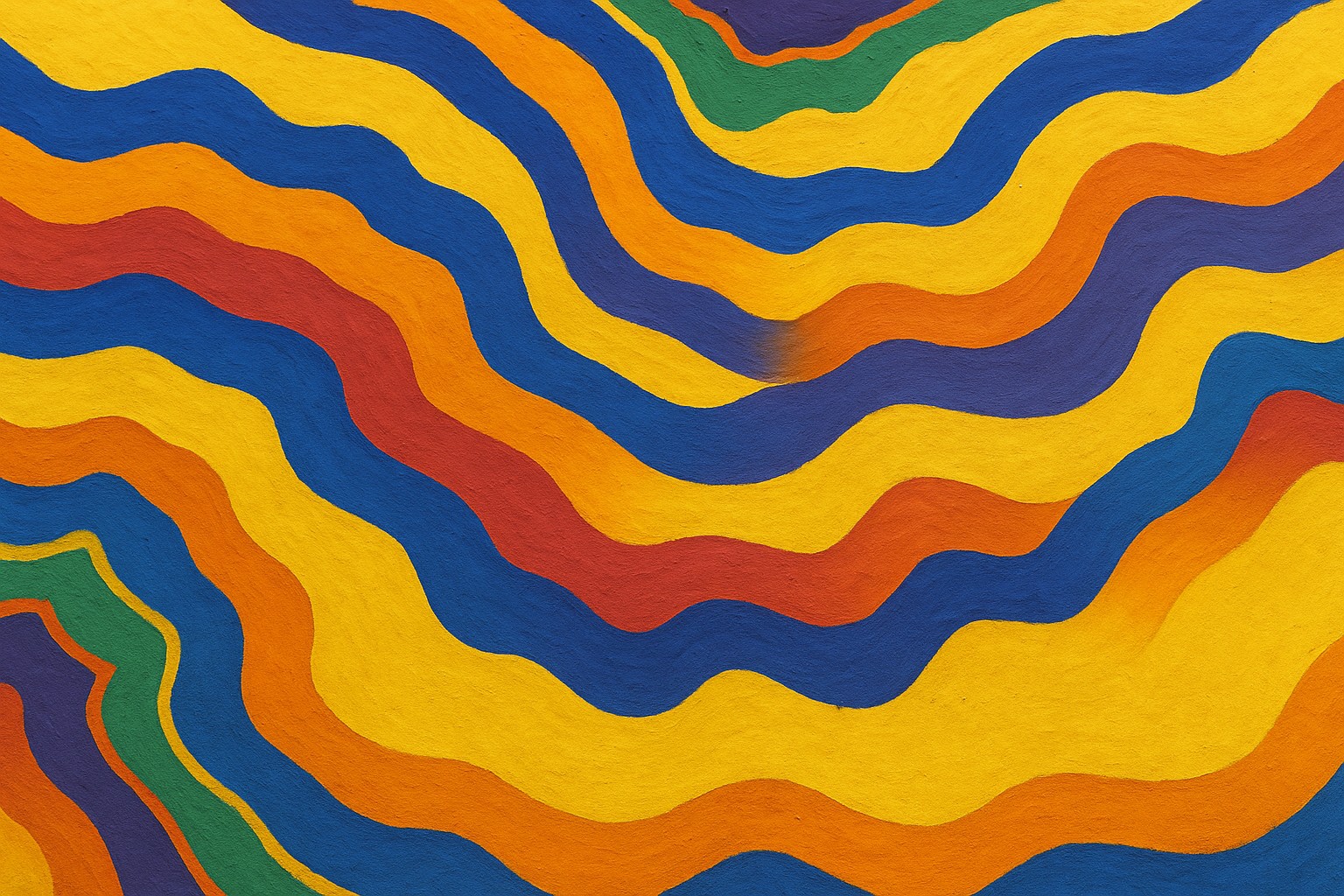}
        \caption*{(iii) Art 1}
    \end{subfigure}
    \quad
    \begin{subfigure}[c]{0.3\textwidth}
        \includegraphics[width=\linewidth]{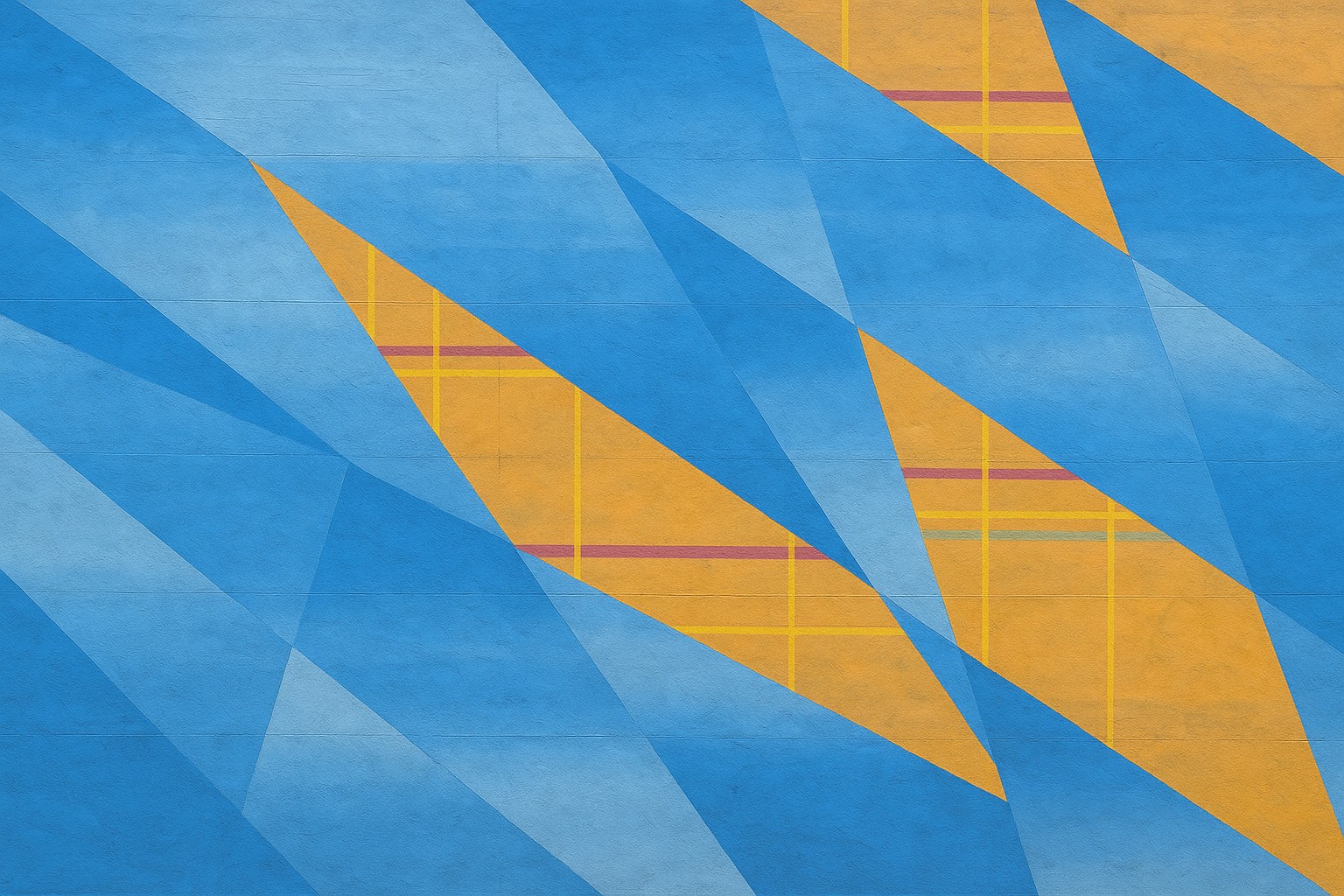}
        \caption*{(iii) Art 2}
    \end{subfigure}

    \vspace{8pt}

    \begin{subfigure}[c]{0.3\textwidth}
        \includegraphics[width=\linewidth]{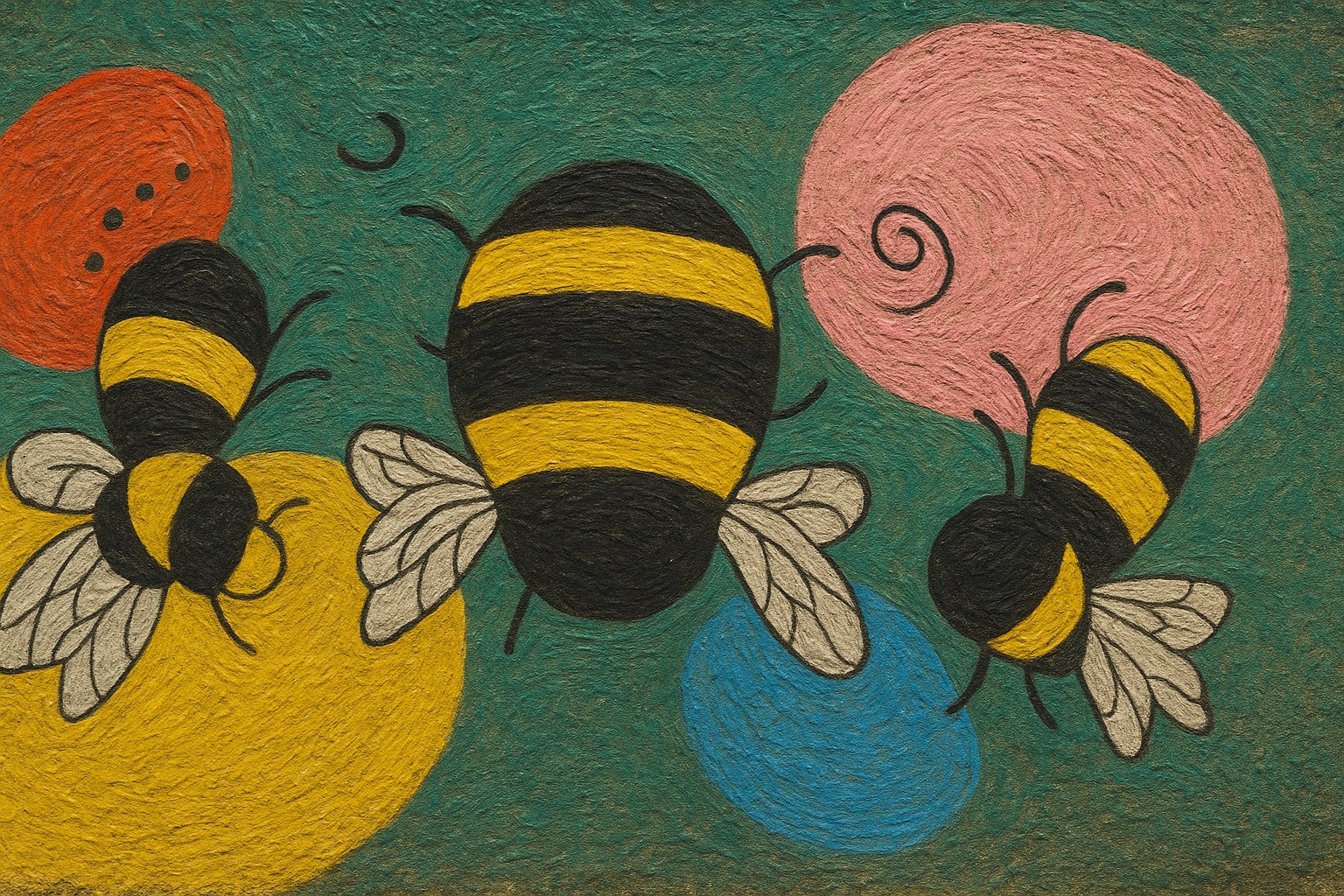}
        \caption*{(iv) Bees}
    \end{subfigure}
    \quad
    \begin{subfigure}[c]{0.3\textwidth}
        \includegraphics[width=\linewidth]{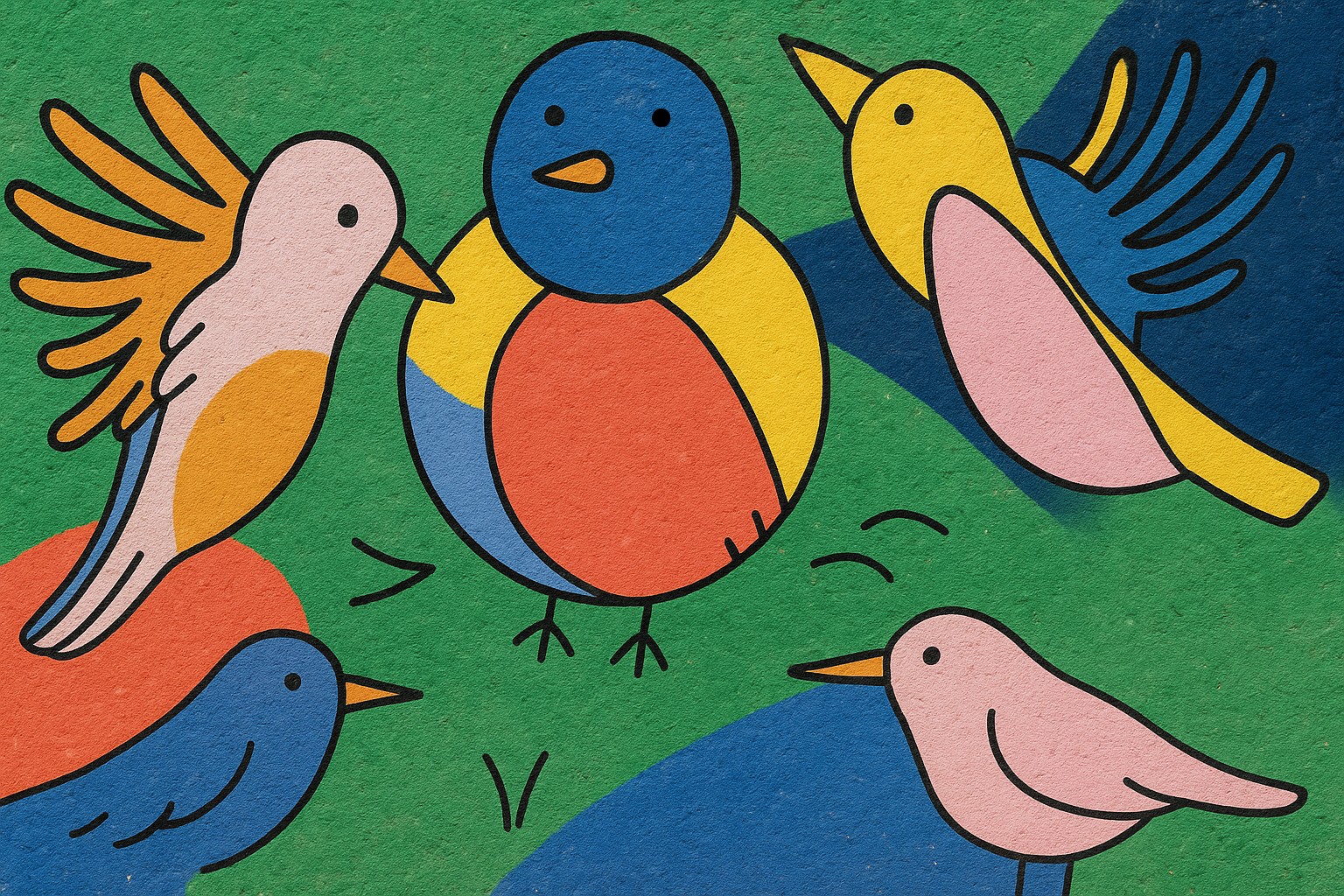}
        \caption*{(iv) Birds}
    \end{subfigure}
    \quad
    \begin{subfigure}[c]{0.3\textwidth}
        \includegraphics[width=\linewidth]{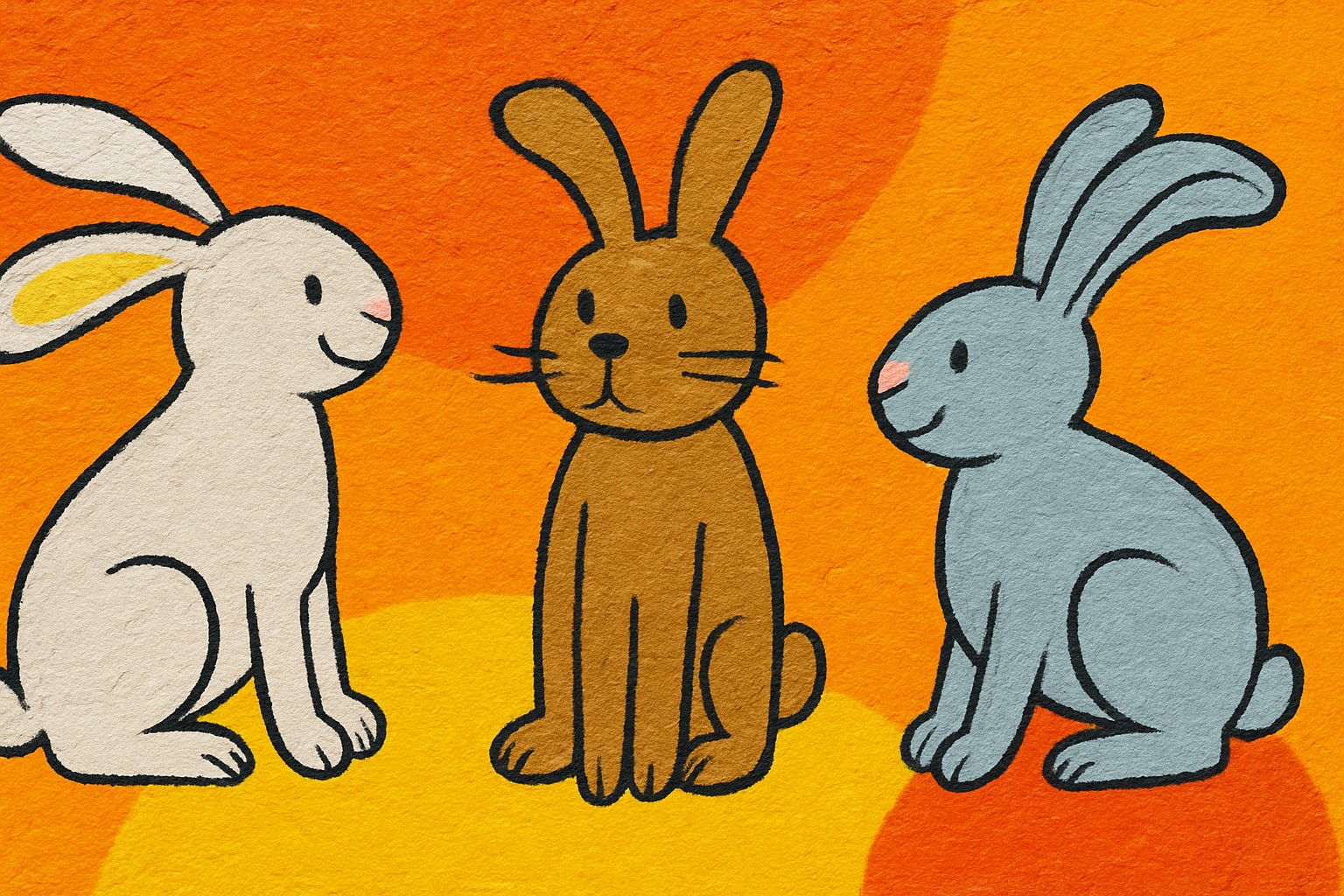}
        \caption*{(iv) Rabbits}
    \end{subfigure}

    \caption{~Examples of asphalt art patterns used in the study, (i)-(iv) denote their categories.}
    \label{fig:asphalt_art_gallery}
\end{figure}

\textbf{2. Asphalt art selection.} To capture a broad spectrum of street art designs, we curated nine representative images from the Bloomberg Philanthropies Asphalt Art Initiative~\cite{bloomberg}. Since the original crawled images depict real-world painted installations, we extracted the artwork patterns from the underlying street textures using a custom workflow assisted by ChatGPT’s vision capabilities~\cite{chatgpt}. This enabled us to get the exact arts such that we can apply them to our collected dataset. The final extracted asphalt arts are shown in Figure~\ref{fig:asphalt_art_gallery}. These samples were categorized into four distinct visual types: (i) white stripes on colored backgrounds, (ii) white stripes over patterned backgrounds, (iii) freeform artistic patterns, and (iv) animal-themed paintings. These categories reflect increasing visual difference from traditional crosswalk appearances, ranging from minor aesthetic variations to highly abstract or figurative designs. 

\textbf{3. Perspective warp transformation.} To realistically embed the extracted art into the annotated crosswalk regions, we performed a homographic transformation using OpenCV library~\cite{opencv_library}. For each crosswalk mask, we computed the minimum bounding quadrilateral that tightly encloses the masked region. This quadrilateral serves as the destination for a perspective warp, mapping the rectangular artwork image into the shape and orientation of the real-world crosswalk. This transformation ensures that the art aligns correctly with the camera’s viewpoint and the geometric context of the scene.

\textbf{4. Applying to testing images.} To generate the final images for evaluation, we followed a multi-step compositing procedure. First, we removed foreground objects such as pedestrians and vehicles from the scene using object segmentation method Segment Anything Model 2 (SAM2)~\cite{ravi2024sam2}, applied to the original frames, guided by the manually annotated bounding boxes. Next, the perspective-transformed street art was placed precisely within the masked crosswalk region. Finally, the removed foreground objects were composited back onto the scene, ensuring that the pedestrians walking over the art were restored in their original positions. This process results in photorealistic images that preserve the scene integrity while injecting targeted street art patterns into the background.

\subsection{Malicious Asphalt Art Creation}
The malicious asphalt art creation process closely follows the same pipeline as in the benign asphalt art creation. The key difference lies in Step 4, where, instead of inserting visually appealing and benign street art, we replace the crosswalk regions with adversarial patterns specifically crafted to degrade pedestrian detection performance. In this setting, we consider two distinct types of attacks that a malicious actor may carry out:
\begin{enumerate}
    \item Adversarial Noise Overlay: The attacker perturbs an existing piece of asphalt art by adding carefully optimized adversarial noise to its visual pattern. The resulting image remains perceptually similar to the original art but is adversarially modified to fool the detection model.
    \item Adversarial Art Creation: The attacker designs a completely new crosswalk art pattern from scratch with the sole purpose of misleading the pedestrian detection model.
\end{enumerate}

\subsubsection{Threat Model}
We define a white-box threat model in which a malicious attacker seeks to compromise the performance of a surveillance-based pedestrian perception system. The attacker has full knowledge of the victim model architecture YOLOv7, and leverages this information to craft visual perturbations optimized to reduce detection performance. The attacker can either modify existing asphalt artwork already present on the street (\eg, by overlaying adversarial noise), or deploy a newly generated asphalt art pattern designed entirely for adversarial purposes. Importantly, the attacker does not alter the pedestrians themselves, nor add any synthetic foreground elements. All modifications are restricted to the crosswalk regions, making the attack less conspicuous and more practical in real-world urban environments.

\subsubsection{Adversarial Noise Overlay}
For the adversarial noise overlay approach, we assume an attacker can train a single perturbation $\Delta$ for each asphalt art pattern and add it to the original artistic crosswalk image. The attack is optimized to degrade YOLOv7’s detection performance, following the methodology of Thys \etal~\cite{thys2019fooling}. Let the set of $N$ object proposals produced by YOLOv7 be $\{P_i\}_{i=1}^N$, where each proposal $P_i = \bigl[x_i,\,y_i,\,w_i,\,h_i,\,\text{obj}_i,\,\text{cls}_i^1, \dots, \text{cls}_i^M\bigr]$. Here $(x_i,y_i,w_i,h_i)$ are the coordinates and dimensions of the bounding box, $\text{obj}_i$ is the objectness score (the probability that $P_i$ contains any object), $\text{cls}_i^j$ is the confidence score for class~$j$, for $j=1,\dots,M$. In our pedestrian‐only setting, $M=1$ and $\text{cls}_i^1 = 1$ for all proposals, thus the class confidence score does not help the optimization process. We define the adversarial objective to suppress YOLOv7’s objectness scores across all proposals.  Concretely, we minimize the loss function $\mathcal{L}_{\mathrm{obj}}$:
\hfill\break
\begin{equation}
  \mathcal{L}_{\mathrm{obj}}(\Delta) = \max_{1 \le i \le N}\sigma\bigl(\text{obj}_i(I+\Delta)\bigr),
  \label{eq:obj_loss}
\end{equation}
\hfill\break
where $\Delta$ is the adversarial noise added to the crosswalk image, $\text{obj}_i(I+\Delta)$ is the objectness logit of the $i$-th proposal after applying $\Delta$ to benign asphalt art image $I$, $\sigma(\cdot)$ denotes the sigmoid activation. By minimizing \(\mathcal{L}_{\mathrm{obj}}\), we drive down the objectness scores of all proposals below the detection threshold, aiming to cause YOLOv7 to miss pedestrians.

\subsubsection{Adversarial Art Creation}
In this adversarial‐art scenario, we assume the attacker has full freedom to craft and apply custom patterns onto crosswalk surfaces, and leverages ChatGPT’s image‐generation feature as a rapid prototyping tool. The attacker works in a heuristic process based on visual insights, rather than a formal optimization pipeline. 
By analyzing data from prior benign trials, the attacker steers the image‐generation process toward increasingly potent perturbations. The resulting adversarial artwork is optimized to achieve two simultaneous objectives: (1) rendering a real pedestrian standing on the crosswalk undetectable; (2) inducing false positive detections of a non‐existent pedestrian purely from the crafted pattern. Our goal is to implement both objectives within a single, highly effective malicious art design.
\section{Results and Analysis}
We measure the YOLOv7’s performance in detecting pedestrians across the created data using standard object detection metrics: max-F1, precision (@max-F1), recall (@max-F1), mAP (mean average precision), and mAP@0.5. These metrics provide a comprehensive view of detection quality: the max-F1 score is the highest F1 value achieved by sweeping the model’s confidence threshold. Precision (@max-F1) reflects the fraction of true pedestrians among all detections when operating at the threshold that yields the max-F1 score. Recall (@max-F1) measures the fraction of ground-truth pedestrians that the model successfully detects at that same threshold. mAP is the standard metric for quantifying a model’s overall detection accuracy in object detection. For a given class, the Average Precision (AP) score is computed from the area under the precision–recall curve, which reflects the trade-off between correctly identifying objects and avoiding false detections. The mAP score is then obtained by averaging AP values across all object classes in the dataset; in our case, this reduces to a single AP value because the dataset contains only one class, "person". Following common evaluation protocols, we report mAP over a range of Intersection-over-Union (IoU) thresholds from 0.5 to 0.95 in increments of 0.05, providing a comprehensive view of detection quality under increasingly strict localization criteria. We also include the widely used mAP@0.5, a special-case metric computed at an IoU threshold of 0.5, which emphasizes correct object identification with more relaxed requirements.

In addition, to capture how these asphalt art patterns induce erroneous detections by YOLOv7, we present the False Discovery Rate (FDR) as a complementary metric, where FDR = FP / (FP+TP). Here, a true positive (TP) is any predicted bounding box with IoU > 0.5 against a ground-truth box and confidence > 0.3, and a false positive (FP) is any prediction above that confidence threshold which fails the IoU criterion. Note that, unlike the other metrics, where lower values denote poorer detection performance, a higher FDR indicates worse behavior by YOLOv7. By combining these metrics across different evaluation settings, we quantify the effect of crosswalk art on the surveillance-based pedestrian perception system.

The YOLOv7 model was trained using the training split (360 images) of our collected dataset, ensuring it learned to detect pedestrians under standard visual conditions. All simulations, both benign and adversarial, and their corresponding evaluations were conducted exclusively on the held-out testing set (77 images) to maintain a clear separation between training and evaluation phases. The 77 testing images were randomly selected from the collected dataset. We analyzed the test set and confirmed that it captures diverse pedestrian scenarios, including variations in pedestrian appearance, pose, count, and spatial distribution. Importantly, all pedestrians present in the selected images were retained, ensuring that the inherent variability of the original dataset is preserved. The number of pedestrians per image ranges from 1 to 4. The test images include pedestrians across all three crosswalks at the T-intersection, providing broad coverage of the different pedestrian scenarios observed across all data collection sessions.

\subsection{Model Evaluation under Clean Scenario}
The baseline model was trained using the standard YOLOv7 augmentation pipeline defined in \cite{wang2023yolov7}. This includes HSV color jitter (hue = 0.015, saturation = 0.7, value = 0.4), random translation (0.2), scaling (0.9), random horizontal flips (p = 0.5), mosaic augmentation (p = 1.0), and mixup (p = 0.15). These augmentations substantially alter both the spatial layout and visual appearance of images, effectively modifying the background, pedestrian placement, and lighting statistics at each iteration. As a result, the model is not exposed to a fixed or static background during training, reducing the risk of overfitting to the specific scene captured in the raw footage. The model was trained for 300 epochs, with no early stopping applied. A batch size of 16 was used, and the best-performing checkpoint was selected based on the highest mAP on the validation set. YOLOv7’s default random-seed initialization was used, which sets a consistent base seed (equivalent to 1 in a single-GPU setup) to ensure reproducible data shuffling and augmentation behavior across training runs.

We evaluate the model using the testing data from the clean scenario (non--asphalt art, regular test data), where it achieves max-F1 $=0.995$, recall $=1.0$, and mAP@0.5 $=0.999$, as presented in Table~\ref{tab:benign}. To ensure that the model is not overfit to a static scene and to demonstrate that its performance remains high when the background changes slightly, we performed further analysis. We trained the model on three of the four sessions and tested it on the remaining session. Across the four train/test combinations, the recall and mAP@0.5 values were: $0.983$ and $0.96$; $0.98$ and $0.99$; $1.0$ and $0.998$; and $0.975$ and $0.996$, respectively. These results show that performance remains consistently high across all splits, indicating that the model generalizes well even when the background and scene conditions vary.

\subsection{Benign Asphalt Art Evaluation}
\begin{figure}[ht]
  \centering
  \begin{subfigure}[b]{0.49\textwidth}
    \centering
    \includegraphics[width=\textwidth]{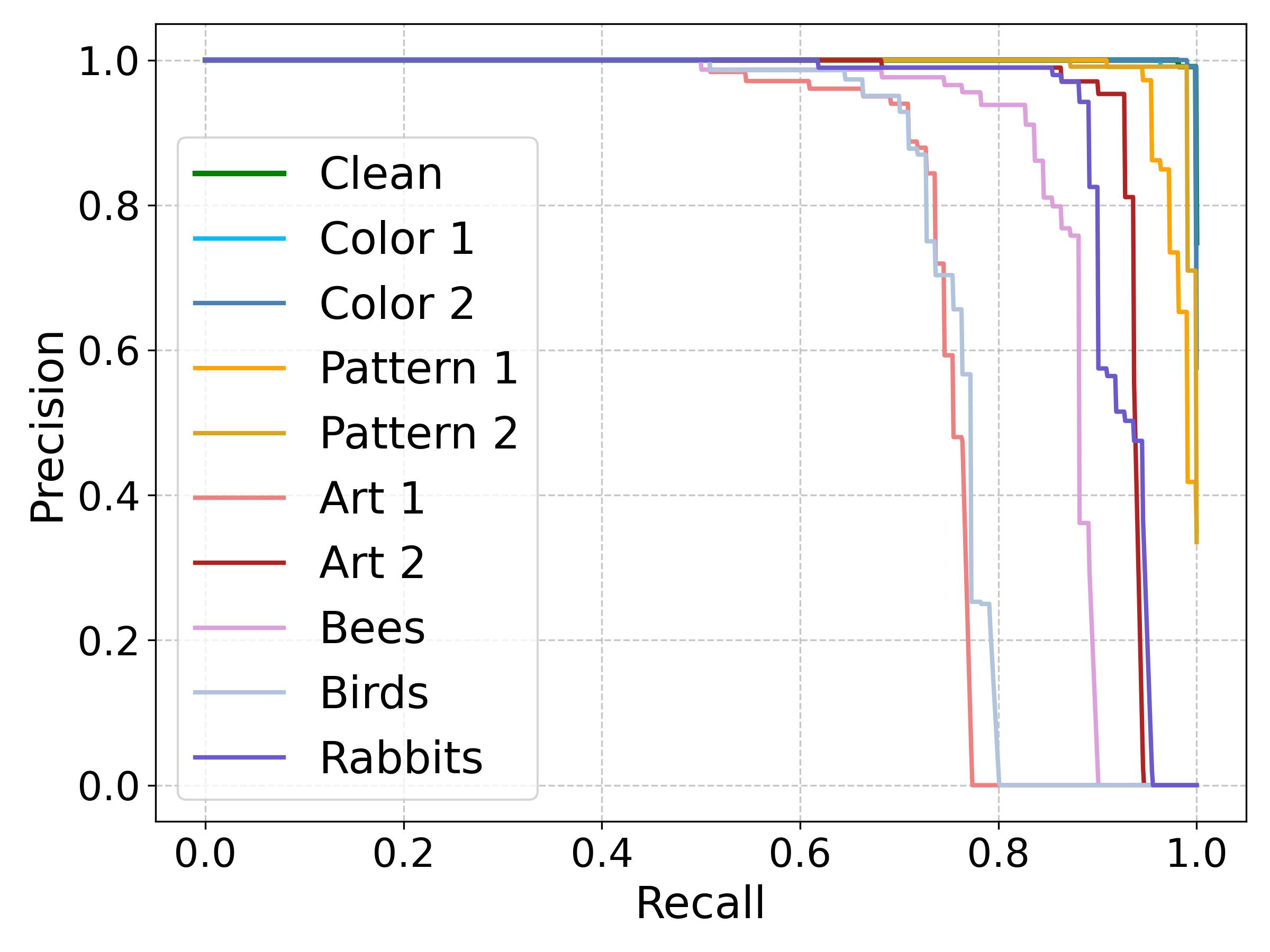}
    \caption{PR curves of benign asphalt art.}
    \label{fig:pr1}
  \end{subfigure}
  \hfill
  \begin{subfigure}[b]{0.49\textwidth}
    \centering
    \includegraphics[width=\textwidth]{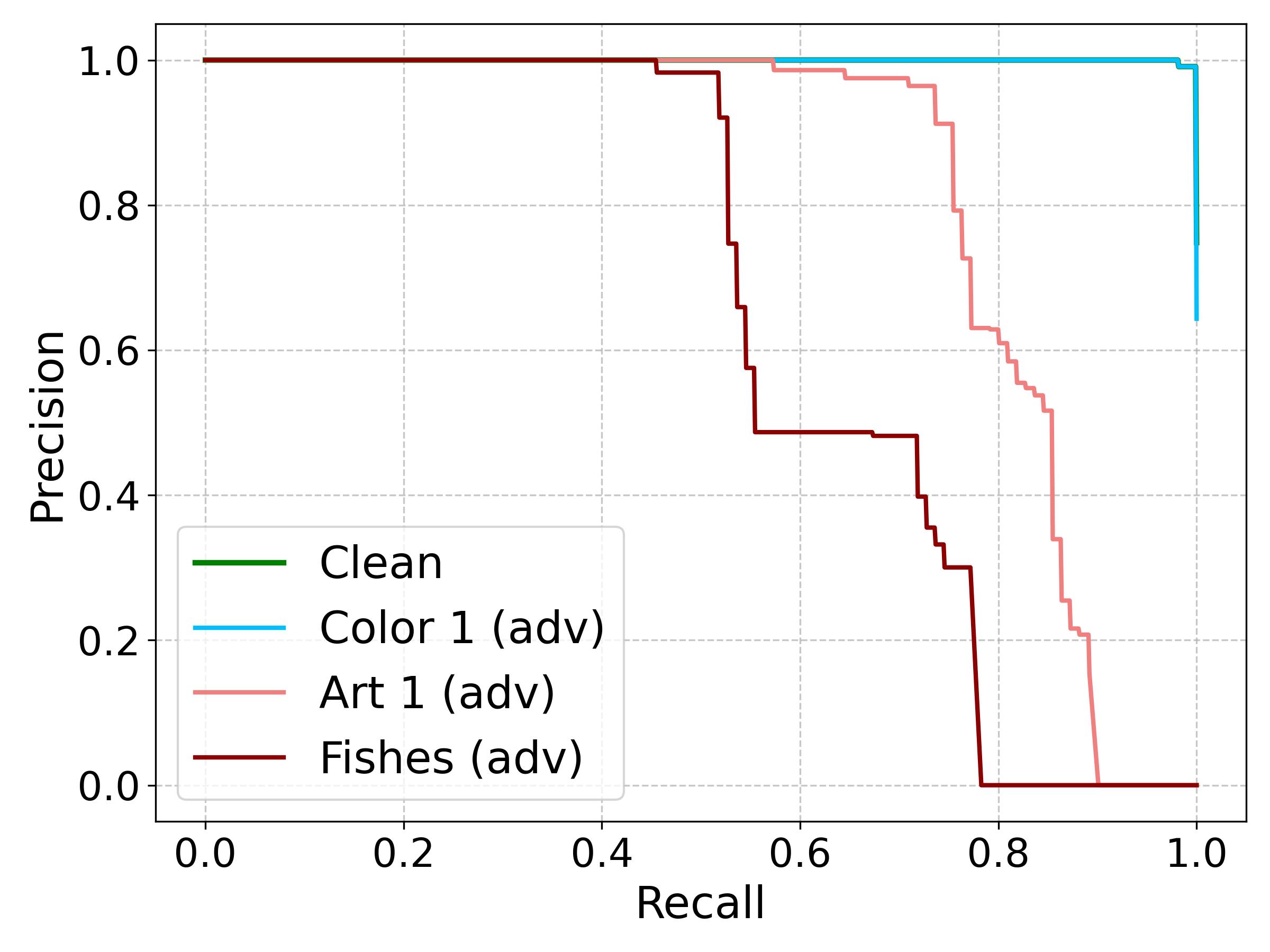}
    \caption{PR curves of malicious asphalt art.}
    \label{fig:pr2}
  \end{subfigure}
  \caption{~PR curves of YOLOv7 with different asphalt art conditions.}
  \label{fig:pr_curves}
\end{figure}

The performance results of the YOLOv7 detector on different benign asphalt art simulations are shown in Table~\ref{tab:benign}. The baseline or clean scenario yields near-perfect results, with a max-F1 of 0.995, a precision of 0.991, a recall of 1.0, an mAP of 0.723, and an mAP@0.5 of 0.999, confirming that YOLOv7 performs reliably under standard visual conditions in the dataset. With a mild artistic background (Color 1 \& 2), YOLOv7 maintains detection performance nearly identical to the clean baseline, with slight drop on mAP. These patterns represent minor visual alterations (white stripes on colored backgrounds) that preserve the crosswalk structure, indicating that YOLOv7 is robust to such aesthetic variations. While moderate artistic backgrounds (Pattern 1 \& 2) lead to minimal degradation, \eg, Pattern 1 exhibits a noticeable drop in recall of 5.5\%, and mAP of 5.9\%. This suggests that more visually complex backgrounds, especially those that interfere with the geometric regularity of crosswalk stripes, begin to affect model reliability. For visually intrusive artworks (Art 1 \& 2, Animals), Art 1 and Birds lead to significant detection failures, with Art 1 dropping to max-F1 of 0.804, mAP of 0.416, and mAP@0.5 of 0.744, significantly lower than the clean baseline. And Birds lead to the lowest recall of 0.700. We show the PR-cures of different art patterns in Figure~\ref{fig:pr_curves} (a), where intrusive artworks (Art 1 \& 2, Animals) generally have lower area under curve than others. These results demonstrate that some asphalt art designs, particularly those with \textit{high visual salience, animal figures, or dense textures}, can significantly interfere with the model’s ability to recognize pedestrians. Furthermore, as shown in our results, all these benign asphalt art patterns have a negligible effect on FDR: even in the worst case (Art 1), FDR increases by only 2.1\% relative to the clean baseline. Our results reveal three key observations:
\begin{itemize}
    \item Increased visual complexity and semantic content in the background art can adversely affect surveillance‐based pedestrian detection, even when the person itself is unaltered.
    \item At the threshold that maximizes F1, precision remains stable, but recall is significantly sacrificed. This phenomenon indicates that, in order to achieve the best F1 performance, many pedestrians are missed due to the disruptive background.
    \item Although these asphalt art patterns lead to missed pedestrian detections, they have minimal impact on false positive predictions.
\end{itemize}

\begin{table}[t]
    \centering
    \caption{~YOLOv7 detection performance on benign asphalt art simulation, with the most impacted performance for each metric highlighted in \textbf{bold}.}
    \begin{tabular}{c|c|c|c|c|c|c}
    \hline
        Arts & max-F1 & Precision & Recall & mAP & mAP@0.5 & FDR \\
        \hline
        Clean & 0.995 & 0.991 & 1.0 & 0.723 & 0.999 & 0.009 \\
        \hline 
        Color 1 & 0.995 & 0.991 & 1.0 & 0.701 & 0.998 & 0.009\\
        Color 2 & 0.995 & 0.991 & 1.0 & 0.699 & 0.998 & 0.009 \\
        Pattern 1 & 0.967 & 0.990 & 0.945 & 0.664 & 0.987  & 0.028 \\
        Pattern 2 & 0.991 & 0.991 & 0.991 &  0.657 & 0.996 & 0.009 \\
        Art 1 & \textbf{0.804} & \textbf{0.927} & 0.709 & \textbf{0.416} & \textbf{0.744} & \textbf{0.030} \\
        Art 2 & 0.938 & 0.953 & 0.923 & 0.581 & 0.933 & 0.011 \\
        Bees & 0.878 & 0.938 & 0.826 & 0.482 & 0.870 & 0.013 \\
        Birds & 0.806 & 0.950 & \textbf{0.700} & 0.457 & 0.756 & 0.014 \\
        Rabbits & 0.924 & 0.970 & 0.882 & 0.510 & 0.919 & 0.012 \\
    \hline
    \end{tabular}
    \label{tab:benign}
\end{table}

\begin{figure}[ht]
  \centering
  \begin{subfigure}[b]{0.48\textwidth}
    \centering
    \includegraphics[width=\textwidth]{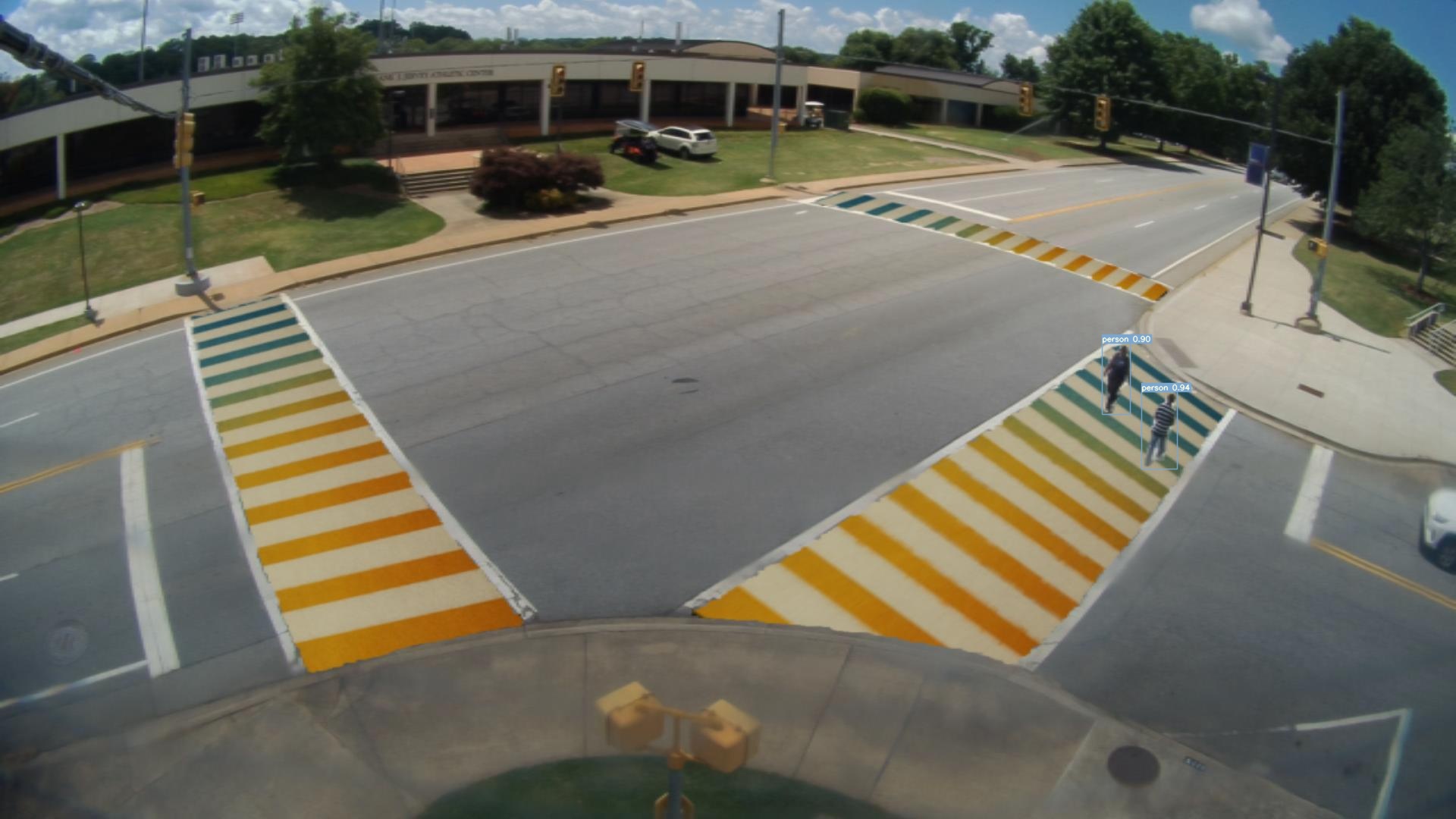}
    \caption{Prediction results of Color 1}
    \label{fig:res1}
  \end{subfigure}
  \hfill
  \begin{subfigure}[b]{0.48\textwidth}
    \centering
    \includegraphics[width=\textwidth]{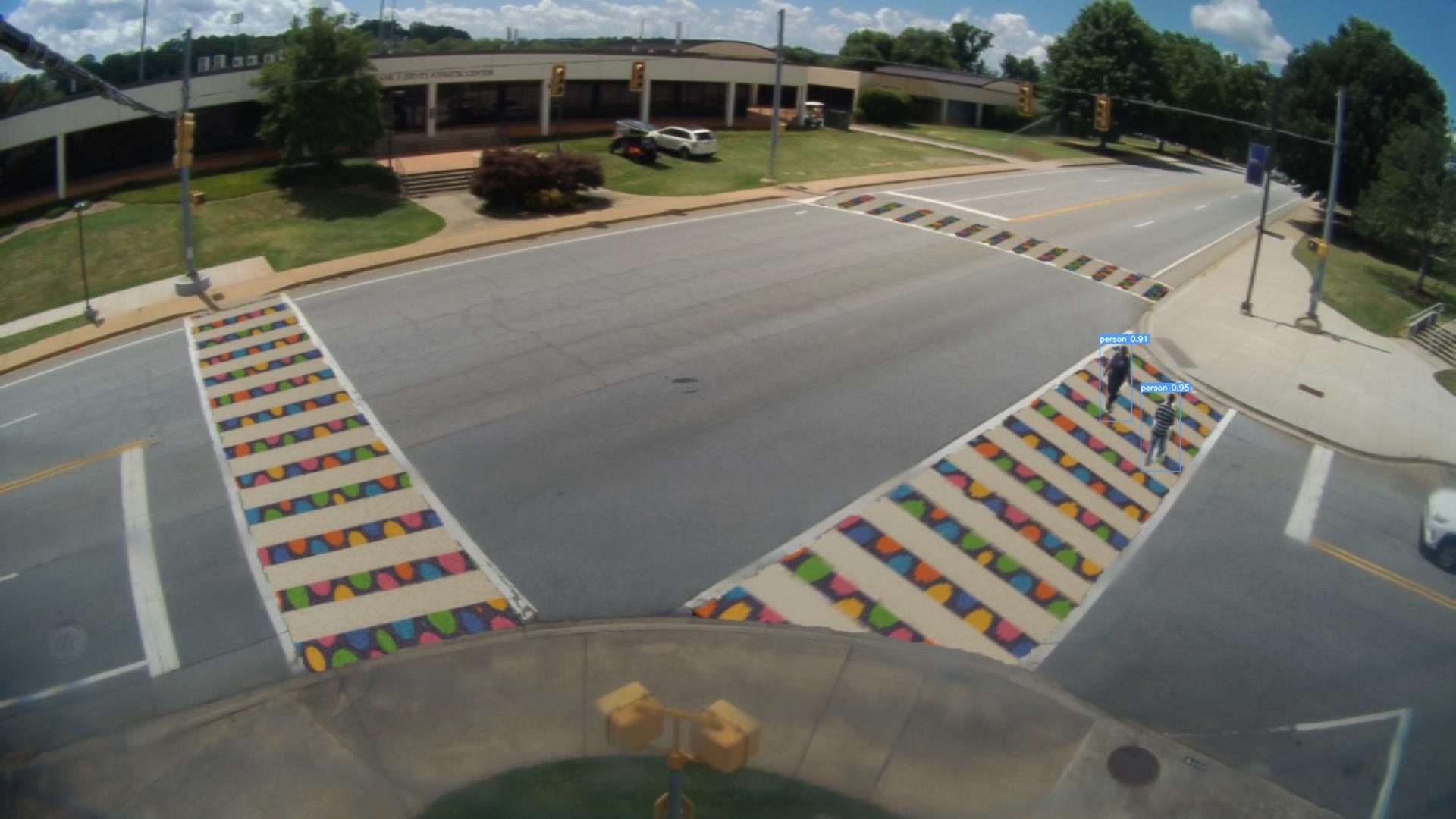}
    \caption{Prediction results of Pattern 1}
    \label{fig:res2}
  \end{subfigure}
  \\
  \begin{subfigure}[b]{0.48\textwidth}
    \centering
    \includegraphics[width=\textwidth]{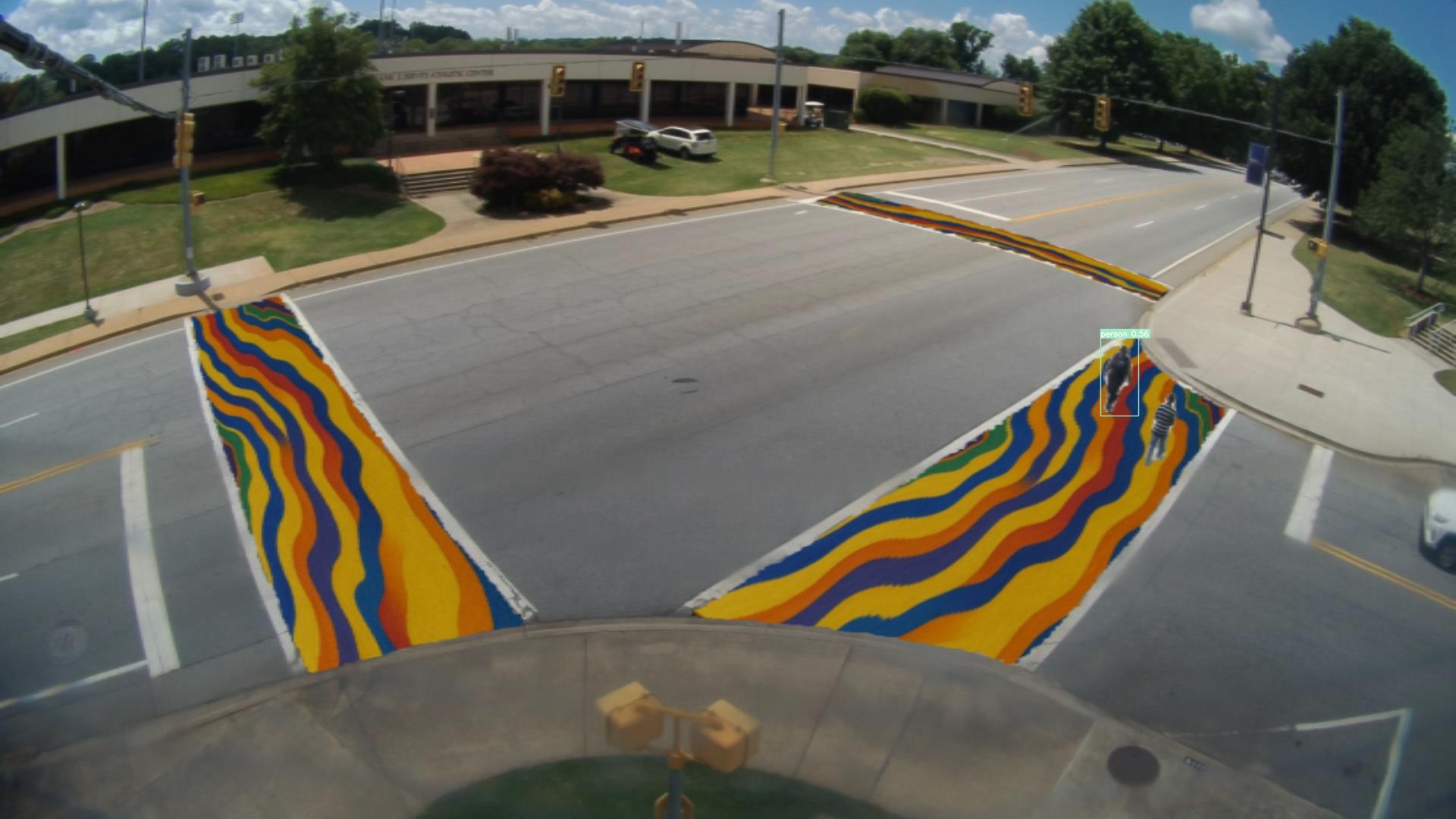}
    \caption{Prediction results of Art 1}
    \label{fig:res3}
  \end{subfigure}
  \hfill
  \begin{subfigure}[b]{0.48\textwidth}
    \centering
    \includegraphics[width=\textwidth]{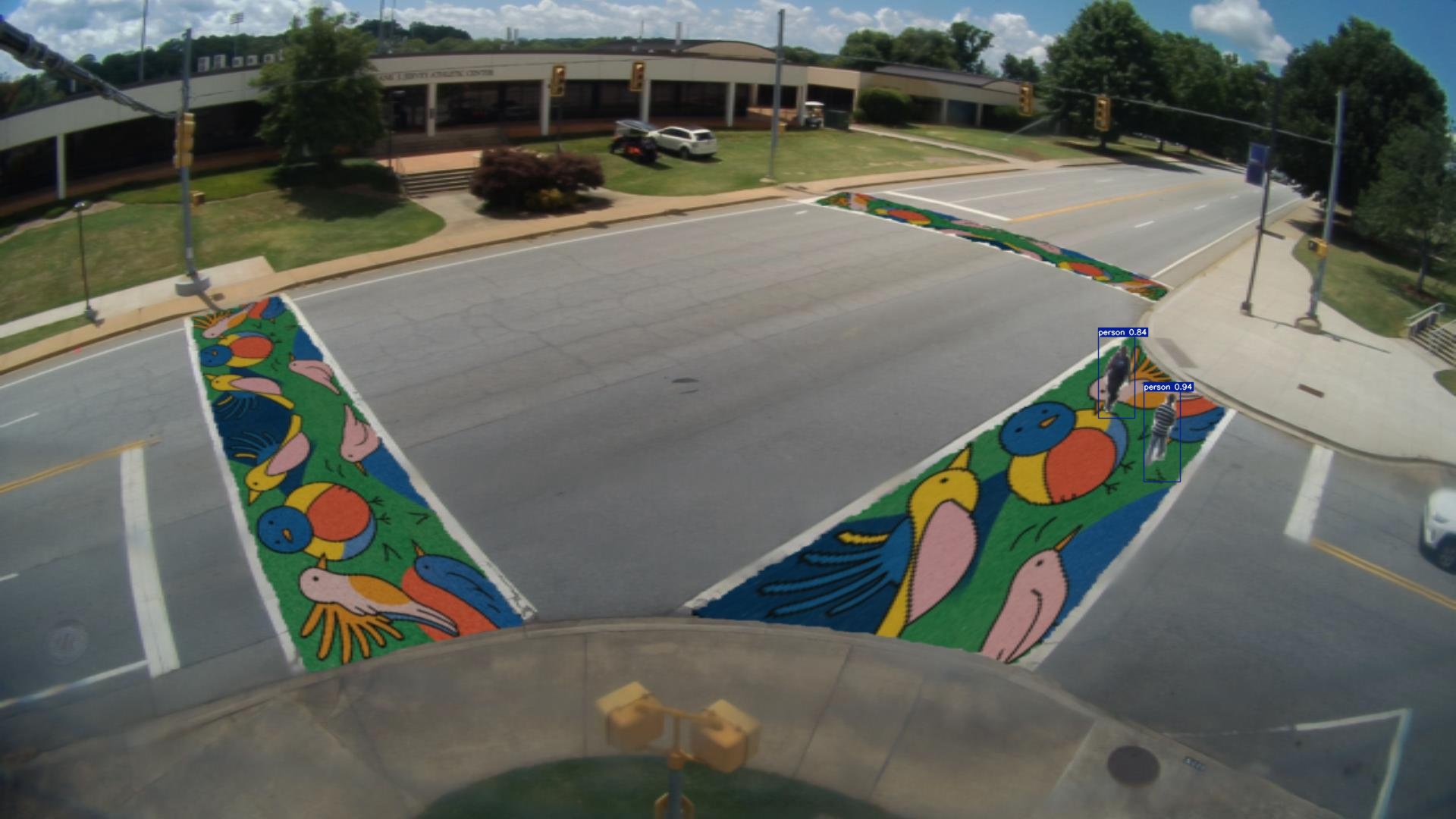}
    \caption{Prediction results of Birds}
    \label{fig:res4}
  \end{subfigure}

  \caption{~Pedestrian detection results comparison of four different asphalt art simulations.}
  \label{fig:qualitative}
\end{figure}

\begin{figure}[t]
  \centering
  \begin{subfigure}[b]{0.245\textwidth}
    \centering
    \includegraphics[width=\textwidth]{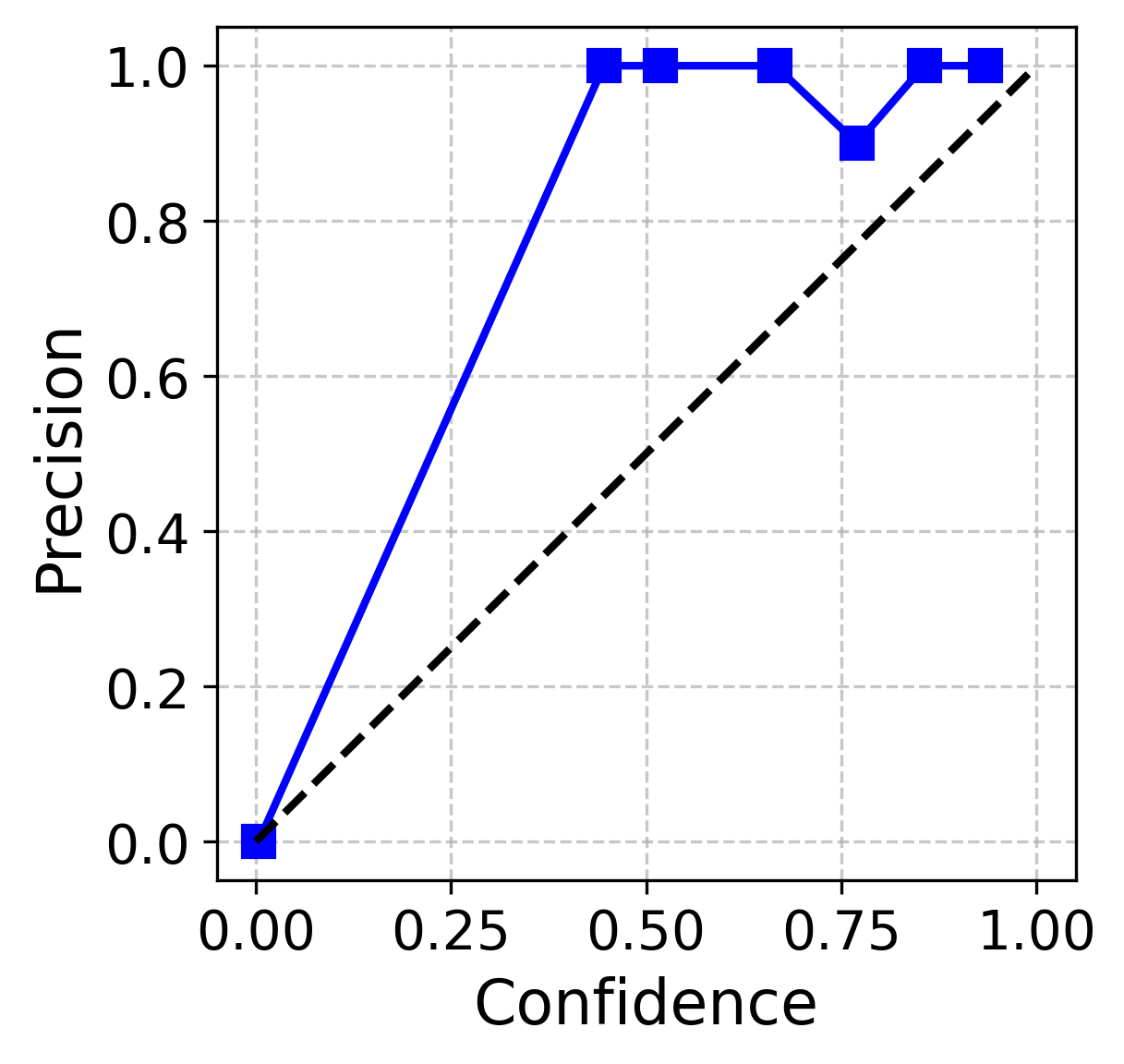}
    \caption{Color 1 (0.014).}
  \end{subfigure}
  \hfill
  \begin{subfigure}[b]{0.245\textwidth}
    \centering
    \includegraphics[width=\textwidth]{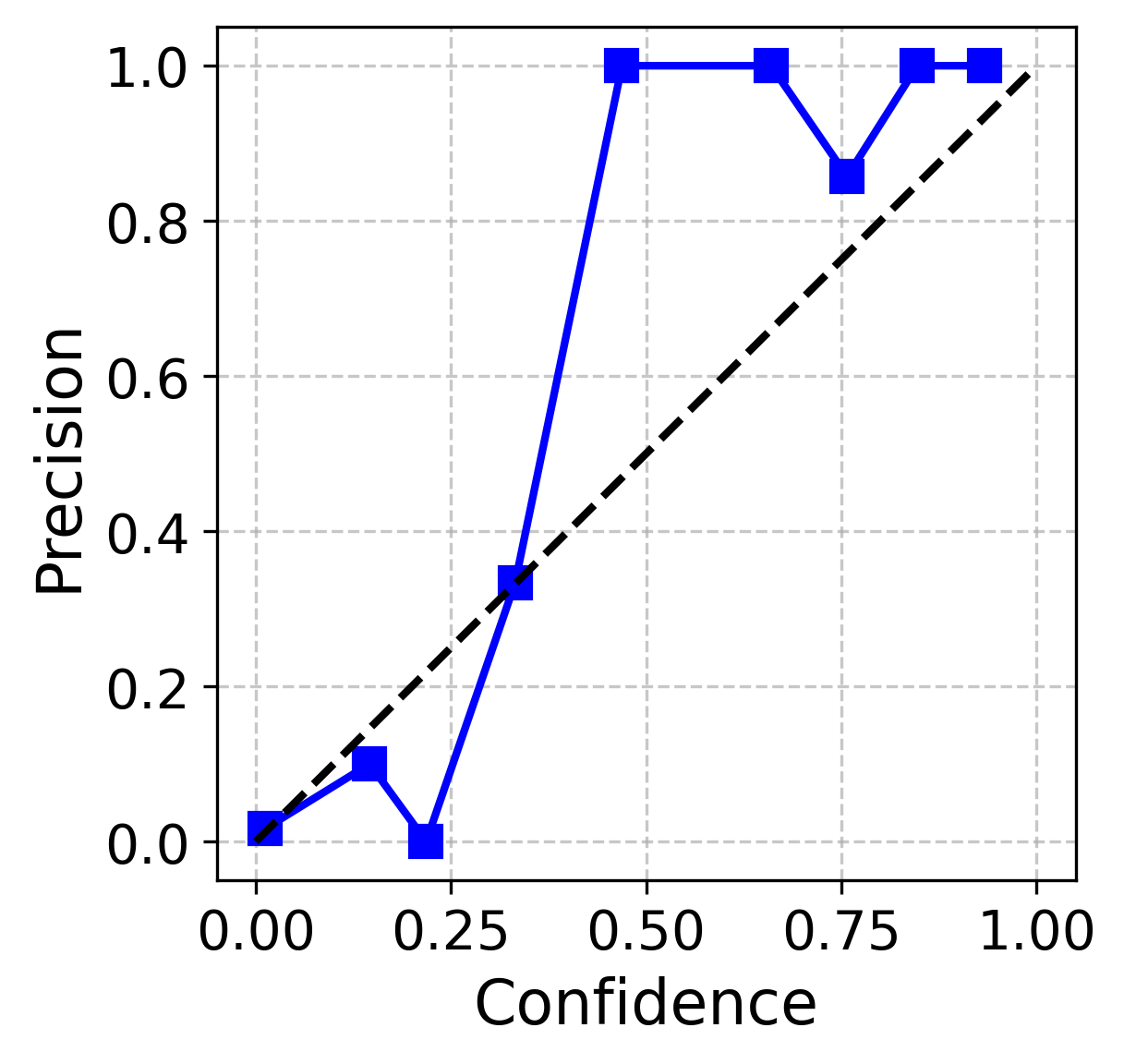}
    \caption{Pattern 1 (0.026).}
  \end{subfigure}
  \hfill
  \begin{subfigure}[b]{0.245\textwidth}
    \centering
    \includegraphics[width=\textwidth]{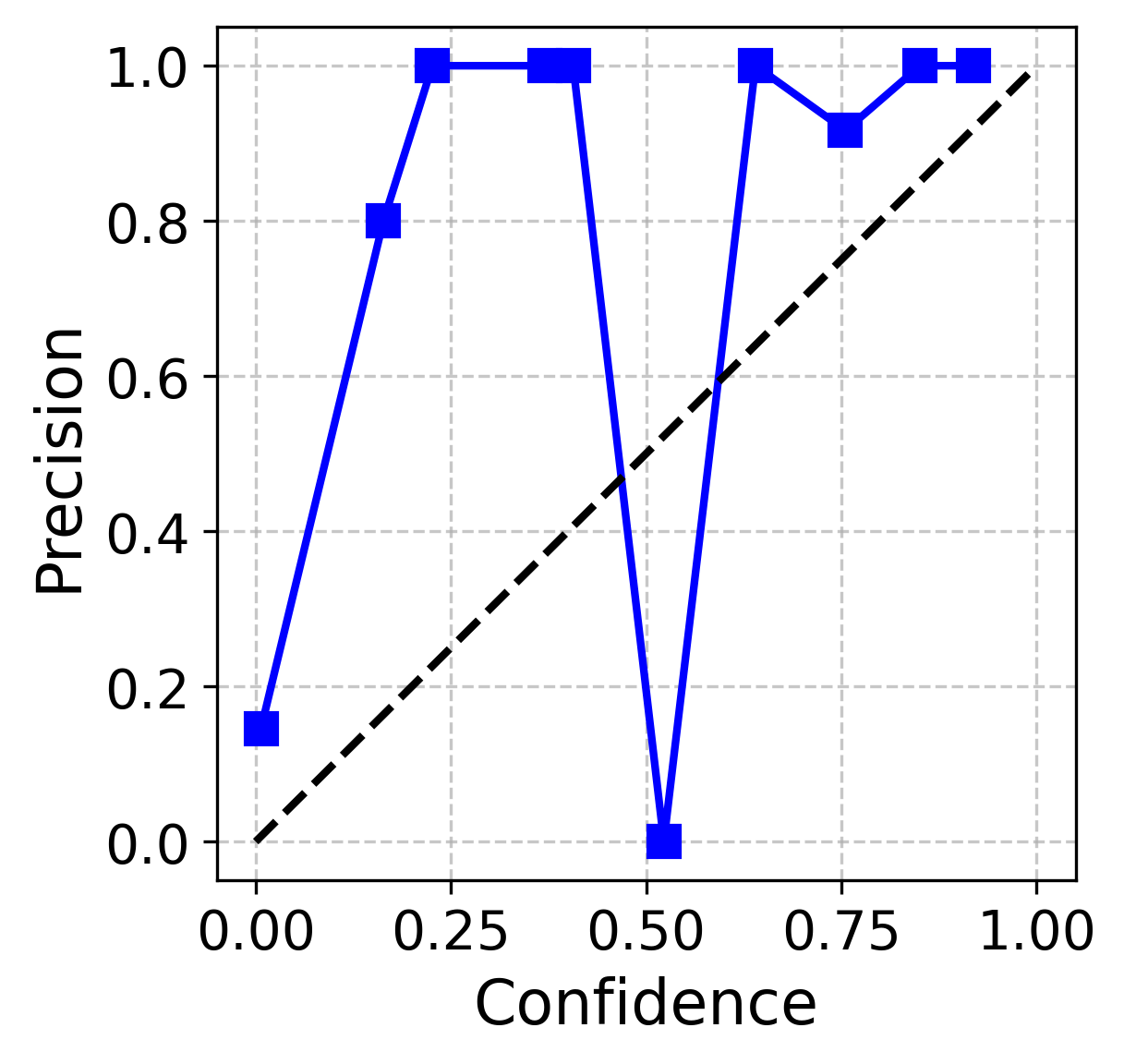}
    \caption{Art 1 (0.120).}
  \end{subfigure}
  \hfill
  \begin{subfigure}[b]{0.245\textwidth}
    \centering
    \includegraphics[width=\textwidth]{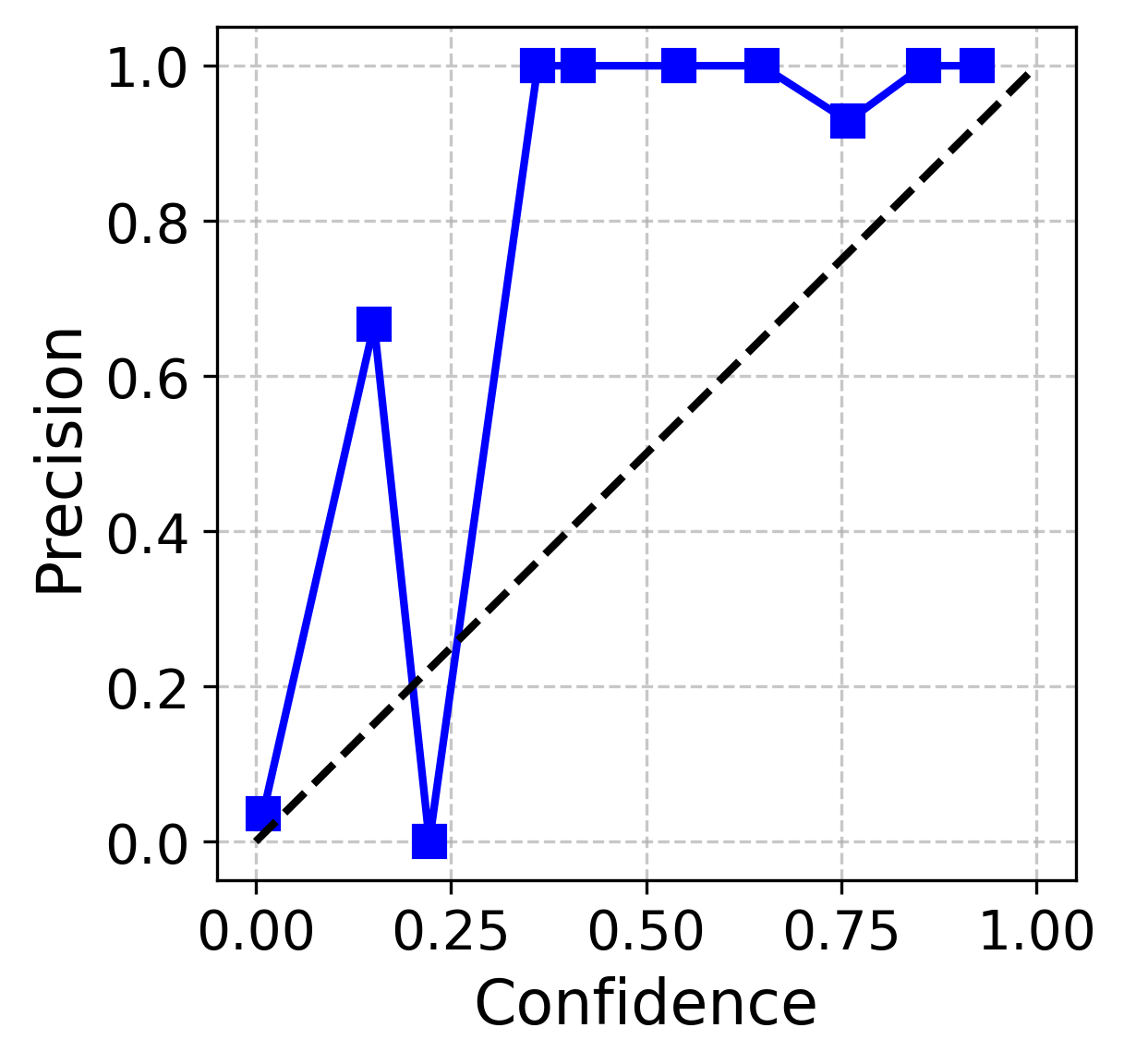}
    \caption{Birds (0.052).}
  \end{subfigure}
  \caption{~Reliability plots of four different benign asphalt art simulations, where the numbers in brackets indicate Brier scores.}
  \label{fig:reliability}
\end{figure}

Figure \ref{fig:qualitative} shows four representative qualitative results, including one example per category (Figure~\ref{fig:asphalt_art_gallery} (i)-(iv)). While YOLOv7 successfully detects the two pedestrians within the image in most scenarios, it fails to identify one of them in the Art 1 case, illustrating its susceptibility to this form of asphalt art.
Figure \ref{fig:reliability} presents reliability plots for the four representative cases, showing how different asphalt art designs influence YOLOv7’s confidence calibration. We also report their corresponding Brier scores in brackets, which quantify calibration quality by measuring the mean squared difference between predicted confidence and actual correctness, where lower values indicate better-calibrated predictions. Across all patterns, the curves generally lie above the diagonal, indicating underconfidence; however, the extent of this effect varies substantially. Color 1 exhibits minimal deviation, reflected by its low Brier score (0.014), suggesting that this simple design has little impact on the model’s reliability. Pattern 1 also shows modest disruption, with a slightly higher Brier score (0.026). In contrast, Art 1 (0.120) and Birds (0.052) introduce noticeably larger calibration errors, with sharper fluctuations and greater divergence from the ideal line. These elevated Brier scores indicate that more visually intricate or semantically rich designs impair YOLOv7’s ability to align its confidence with actual precision, reinforcing the finding that certain artistic patterns can affect model reliability even in the absence of explicit attacks.

\subsection{Malicious Asphalt Art Evaluation}
In the adversarial noise overlay experiments, we optimized two universal perturbations using the loss function in Equation~\ref{eq:obj_loss}: one for the mildest pattern Color 1, the other one for the most disruptive pattern Art 1. These noises were applied exclusively to the crosswalk regions, leaving pedestrians and other scene elements untouched. As Table \ref{tab:malicious} shows, injecting noise into Color 1 has virtually no effect on YOLOv7’s performance, and surprisingly, the addition of noise to Art 1 even yields a slight improvement in detection performance and a slight drop in FDR, indicating fewer false positive detections. We attribute this robustness to YOLOv7’s training strategy, which incorporates extensive data augmentation, such as adding random noise or mosaic, making the model inherently resistant to such kinds of noise perturbations.

For the adversarial art creation approach, building on our earlier finding that \textit{high visual salience, animal figures, and dense textures} most severely degrade detection, we leveraged ChatGPT’s image‐generation capability to produce an adversarial artwork. Since pedestrians occupy only a small footprint in surveillance imagery, we crafted a simple sketch of “fishes” to mimic human shapes, overlaid on a purple-and-orange background, which are the brand colors of Clemson University. As shown in Table~\ref{tab:malicious}, this adversarial art achieves a max-F1 of 0.679, 12.5\% below the worst benign pattern (Art 1); and reduces recall to just 0.518, meaning the detector finds only half of the actual pedestrians. Moreover, the FDR rises to 0.544, 51.4\% higher than Art 1, indicating that over half of YOLOv7's predictions are false positives. Figure~\ref{fig:pr_curves} (b) presents the PR curves corresponding to the different malicious asphalt art designs. The curves for the clean baseline and Color 1 overlap almost entirely, indicating negligible degradation. After adding adversarial noise, Art 1 even becomes less effective in attacking YOLOv7. In contrast, the malicious Fishes pattern produced by our approach yields the smallest area under the curve, demonstrating the strongest reduction in detection performance.

\begin{figure}[t]
    \centering
    \begin{subfigure}[b]{0.44\textwidth}
        \includegraphics[width=\linewidth]{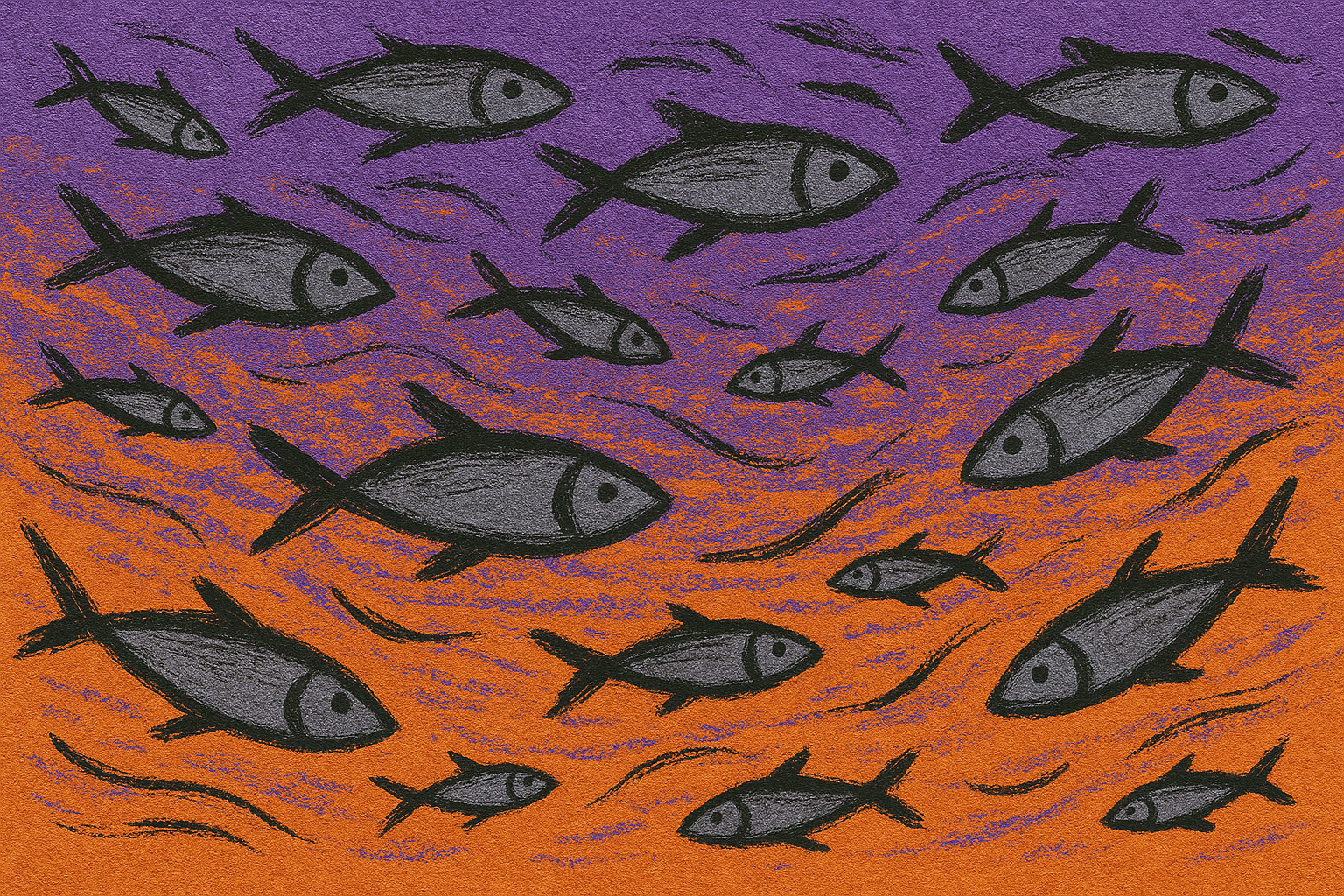}
        \caption{Fishes: final adversarial asphalt art.}
    \end{subfigure}
    \hfill
    \begin{subfigure}[b]{0.52\textwidth}
        \includegraphics[width=\linewidth]{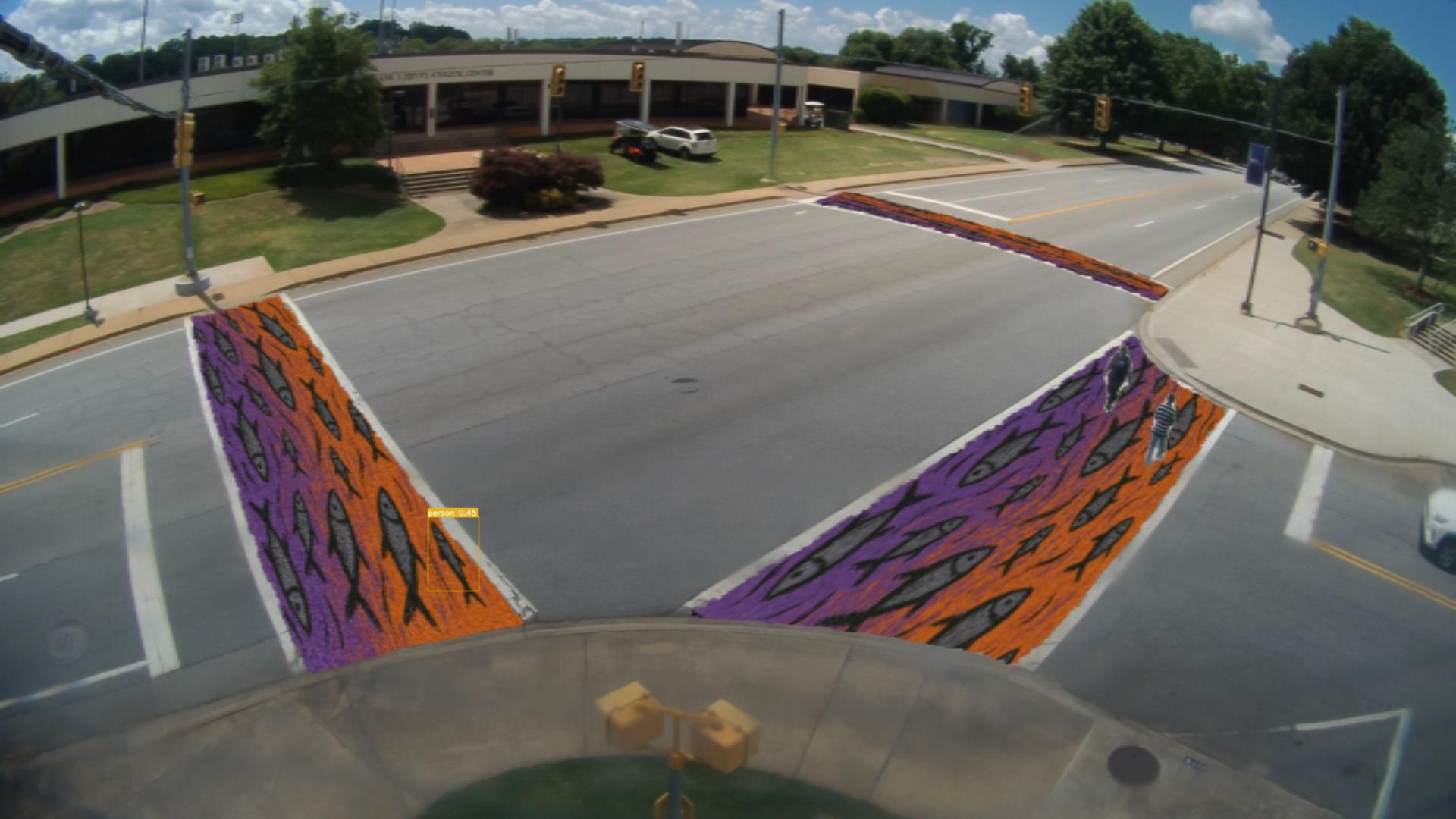}
        \caption{An example image with detection results.}
    \end{subfigure}
    \caption{~The final malicious asphalt art created by the attacker, and one example image with the YOLOv7 detection results.}
    \label{fig:adv_art}
\end{figure}

Figure~\ref{fig:adv_art} (a) presents the final design alongside an example detection output, and Figure~\ref{fig:adv_art} (b) shows one example with detection results. This pattern not only conceals two real pedestrians on the right side, but also induces a false positive detection of a non-existent pedestrian on the left side of the crosswalk. These results demonstrate that carefully designed adversarial artworks can simultaneously execute both obfuscation and hallucination attacks, posing a significant threat to vision-based surveillance systems.

\begin{table}[]
    \centering
    \caption{~YOLOv7 detection performance on malicious asphalt art simulation. The arrows in adversarial noise overlay methods show the trend compared to their benign counterparts.}
    \begin{tabular}{c|c|c|c|c|c|c|c}
    \hline
        Approach & Arts & max-F1 & Precision & Recall & mAP & mAP@0.5 & FDR \\
        \hline
         & Clean & 0.995 & 0.991 & 1.0 & 0.723 & 0.999 & 0.009 \\
        \hline 
        Adv-noise & Color 1 & 0.995$-$ & 0.991$-$ & 1.0$-$ & 0.701$-$ & 0.998$-$ & 0.009$-$ \\
         & Art 1 & 0.835$\uparrow$ & 0.964$\uparrow$ & 0.736$\uparrow$ & 0.482$\uparrow$ & 0.821$\uparrow$ & 0.037$\downarrow$ \\
        \hline
        Adv-art & Fishes & \textbf{0.679} & \textbf{0.983} & \textbf{0.518} & \textbf{0.373} & \textbf{0.641} & \textbf{0.544} \\
        \hline
        
    \end{tabular}
    \label{tab:malicious}
\end{table}

\begin{figure}[ht]
    \centering
    \includegraphics[width=0.8\linewidth]{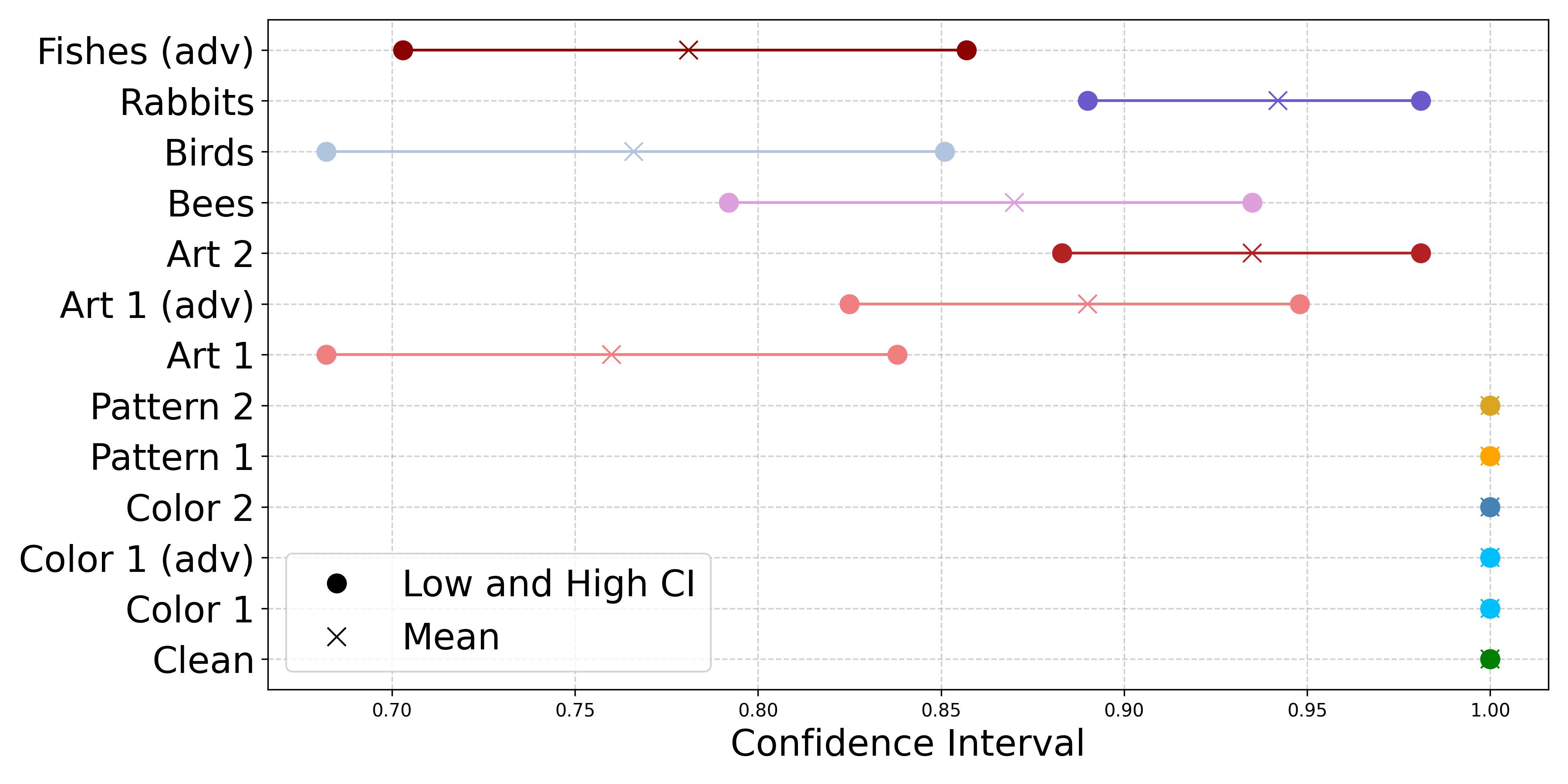}
    \caption{~Bootstrap 95\% confidence intervals (CIs) for image-level recall across all asphalt art conditions.}
    \label{fig:cis}
\end{figure}

To distinguish genuine performance changes from sampling variability, we report uncertainty estimates in the form of bootstrap confidence intervals (CIs) computed across image frames for the recall metric, comparing the clean baseline with different asphalt-art conditions. Figure \ref{fig:cis} summarizes these image-level CIs, providing a direct comparison across all patterns. The clean baseline shows near-perfect recall with an extremely tight CI around 1.0, indicating both high accuracy and low variability. Simple color variations (Color 1, Color 2) and basic geometric patterns (Pattern 1, Pattern 2) similarly yield CIs concentrated near 1.0, suggesting that these benign designs have minimal impact on detection reliability and that YOLOv7 is robust to such simple background changes. In contrast, several visually complex designs, most notably Art 1 and Birds, produce substantially wider and lower CIs, reflecting both reduced recall and greater variability across frames. For the adversarial conditions, Art 1 (adv) shows a somewhat elevated CI relative to its benign counterpart, indicating that this particular noise perturbation may not consistently degrade recall. In contrast, our generated Fishes (adv) pattern exhibits both low mean recall and a wide CI, clearly demonstrating its effectiveness as an adversarial attack. Overall, these results show that while most benign asphalt art designs do not significantly disrupt YOLOv7’s performance, certain intricate or semantically rich patterns can meaningfully degrade detection consistency, and intentionally crafted adversarial artwork can induce severe and reliable failures across frames.

\subsection{Ablation Study}
\subsubsection{Impact of Simulation Pipeline}
We study whether the synthetic simulation pipeline in Figure~\ref{fig:pipeline} introduces any unintended degradation in pedestrian detection. Specifically, we apply the same pipeline to the original test images as an ablation study. Step~1 crosswalk mask annotation remains unchanged. In Steps~2-3, asphalt art selection and perspective warp, instead of inserting asphalt art, we insert the original crosswalk regions. Following our standard procedure in Step~4, we also remove pedestrians and vehicles and then paste them back onto the synthesized images. We find that YOLOv7's detection performance on these control images is largely unchanged: all metrics remain the same, except that mAP drops from 0.723 to 0.698 (2.5\% decrease). This indicates that the simulation process introduces only minor artifacts and largely preserves downstream pedestrian detection performance, while still enabling effective image synthesis.

\begin{figure}[t]
    \centering
    \begin{subfigure}[b]{0.32\textwidth}
        \includegraphics[width=\linewidth]{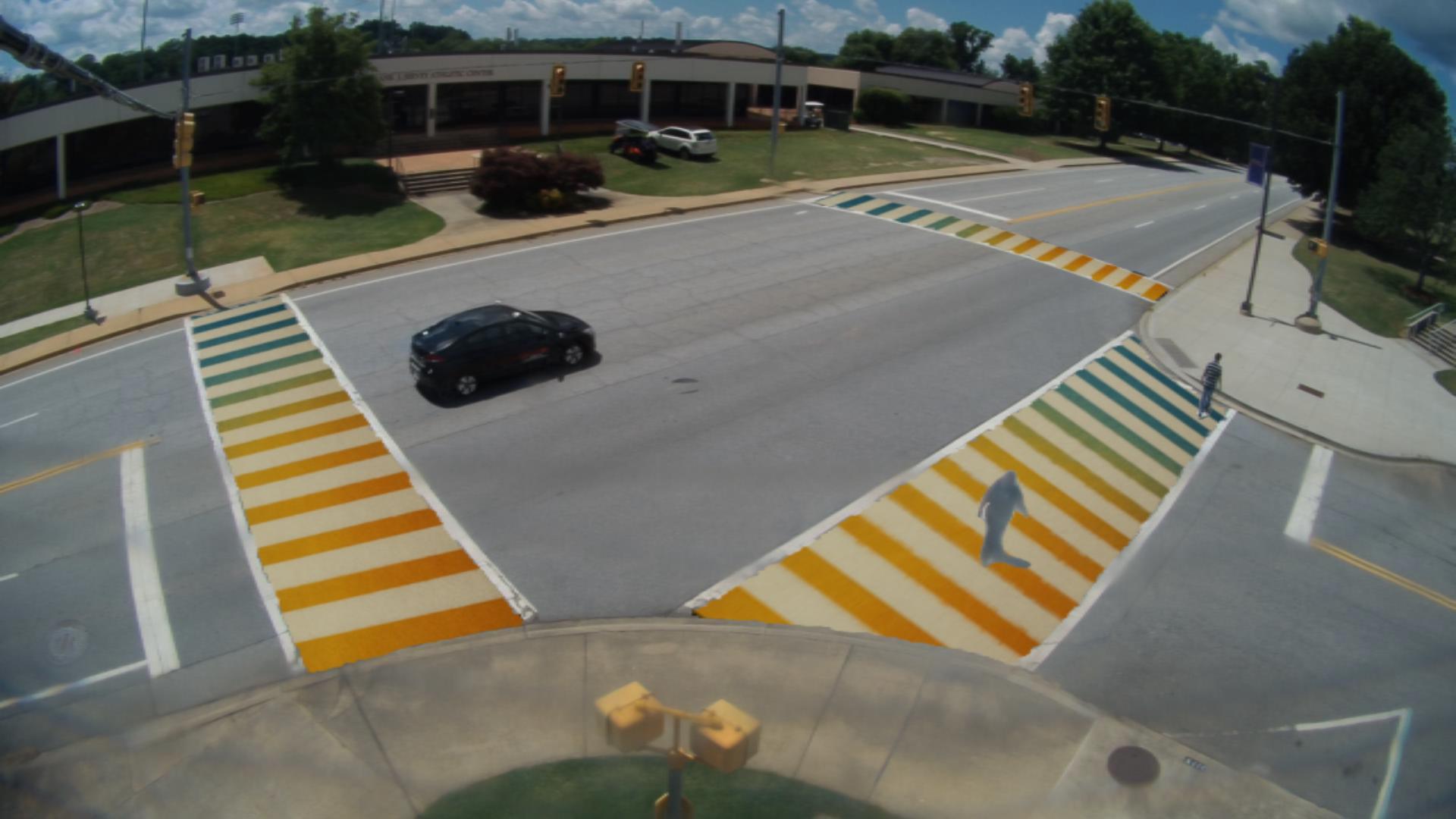}
    \end{subfigure}
    \hfill
    \begin{subfigure}[b]{0.32\textwidth}
        \includegraphics[width=\linewidth]{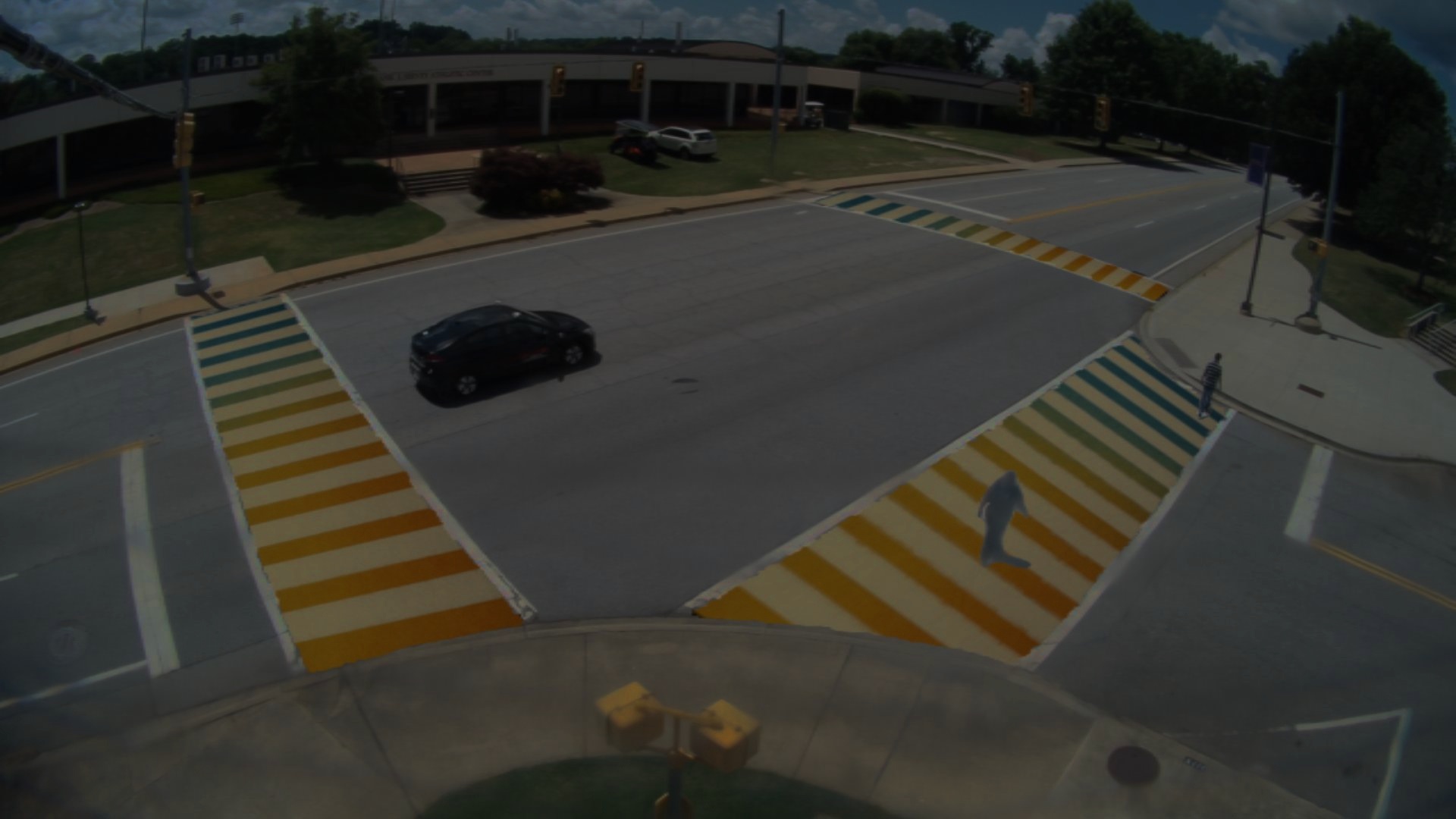}
    \end{subfigure}
    \hfill
    \begin{subfigure}[b]{0.32\textwidth}
        \includegraphics[width=\linewidth]{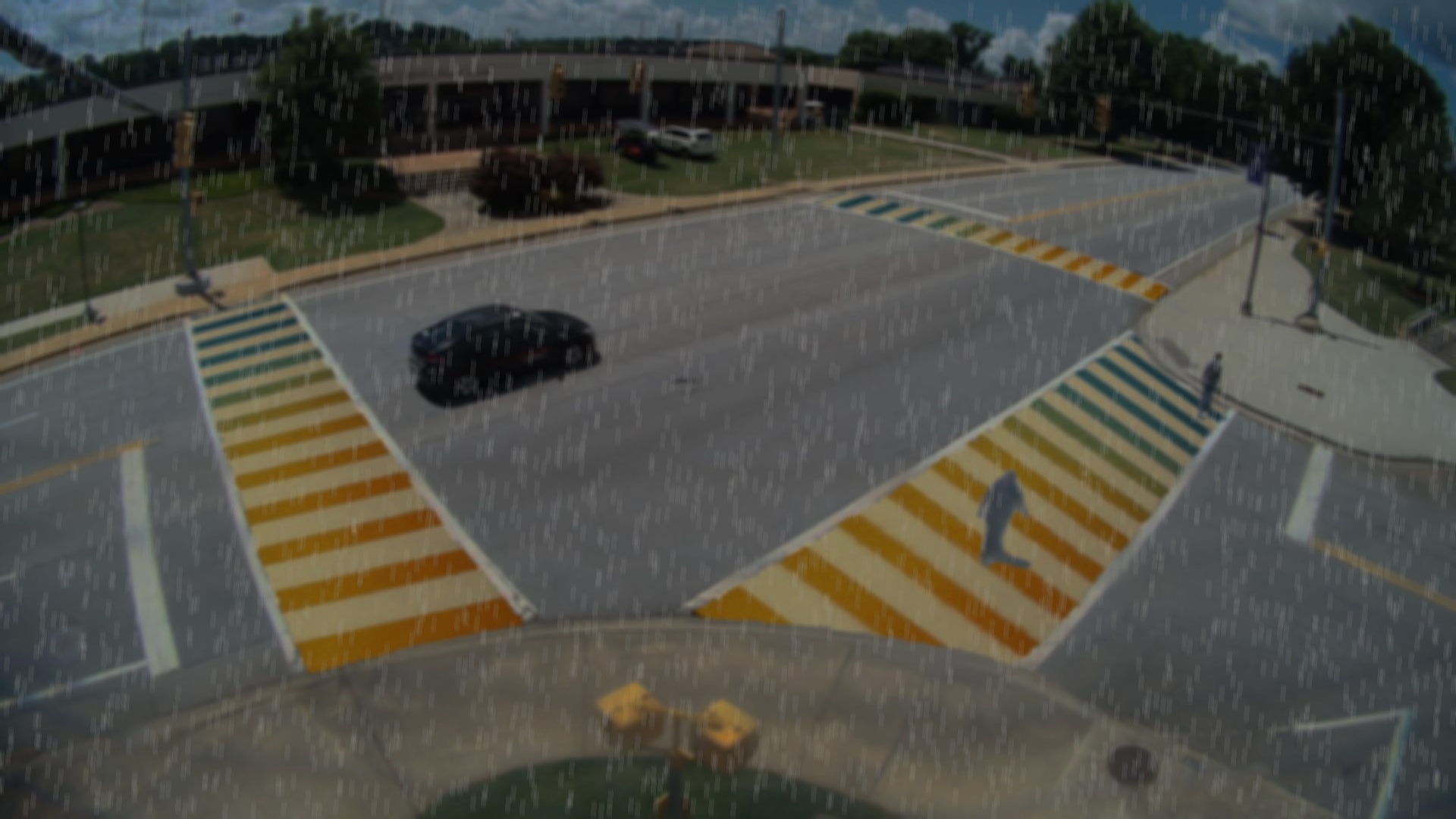}
    \end{subfigure}
    \caption{~Example images after weather-based augmentations with asphalt art Color 1. From left to right: original image, dark weather, rainy weather.}
    \label{fig:weather_examples}
\end{figure}

\begin{figure}[t]
    \centering
    \includegraphics[width=0.8\linewidth]{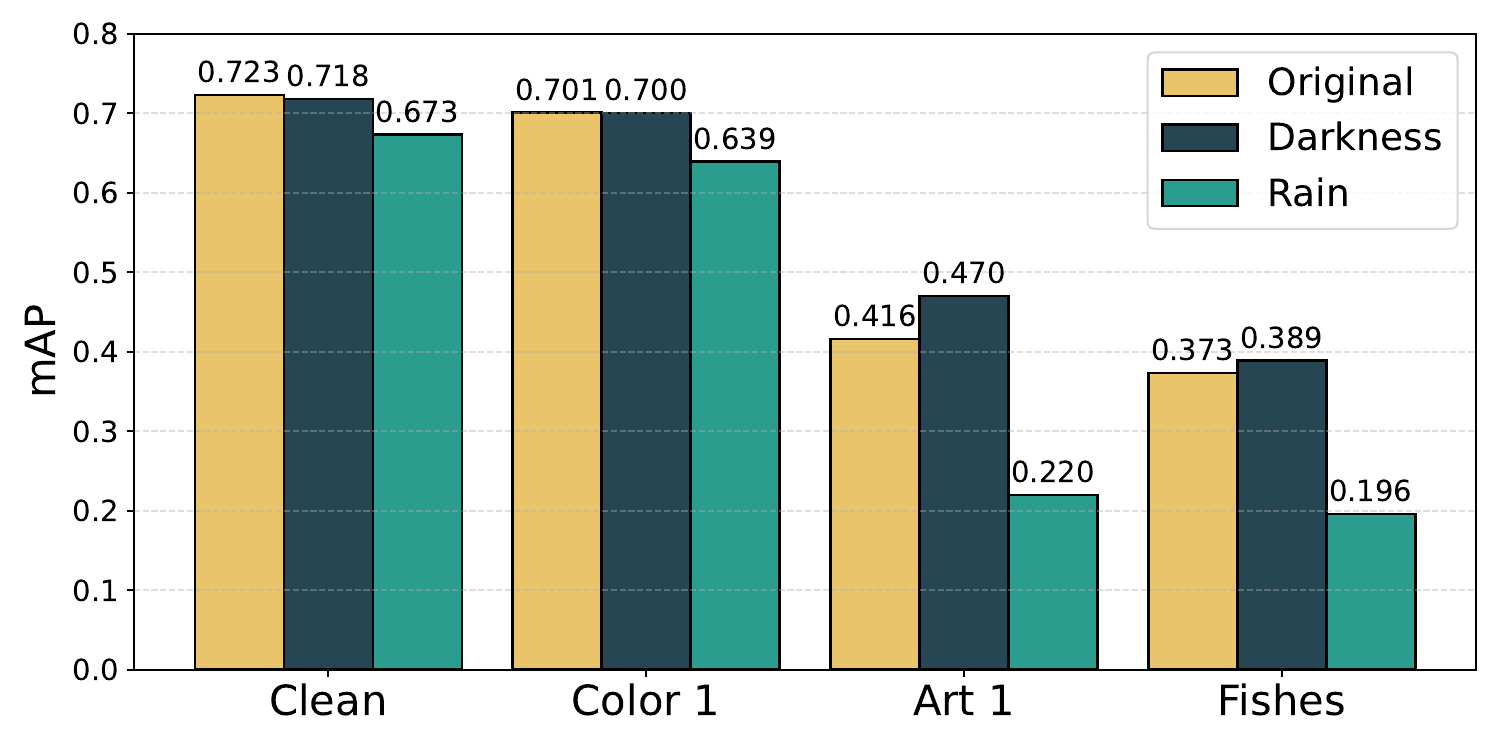}
    \caption{~YOLOv7 mAP performance across four different asphalt art scenarios under darkness and rainy weather conditions.}
    \label{fig:weather_map}
\end{figure}

\subsubsection{Robustness to Different Weather Conditions}
We study the robustness of asphalt art attacks under varying weather conditions as presented in this section. To simulate environmental effects, we use the Automold tool~\cite{automold} to apply weather-based augmentations, specifically darkness and rainy conditions, to the images. These augmentations are applied across four different asphalt art scenarios for comparison: clean images, two benign asphalt art patterns Color 1 (the most moderate one) and Art 1 (the most disruptive one), and the malicious asphalt art, called Fishes, which exhibits the strongest adversarial attack performance on YOLOv7. Figure~\ref{fig:weather_examples} shows example images after the weather augmentations for asphalt art Color 1. In addition, we show the mAP performance of YOLOv7 under different weather conditions in Figure~\ref{fig:weather_map}. For clean images, both darkness and rain cause slight degradation (from 0.723 to 0.718 and 0.673), indicating increased detection difficulty. A similar trend is observed for the moderate benign art Color 1. In contrast, the disruptive benign art Art 1 shows improved mAP under darkness (from 0.416 to 0.470), suggesting reduced attack effectiveness, but a sharp drop under rain (to 0.220), indicating amplified impact. The malicious asphalt art Fishes follows the same trend, with a slight increase of mAP under darkness (from 0.373 to 0.389) and a significant decrease under rain (to 0.196). Overall, asphalt art remains effective across different weather conditions, with rainy weather further exacerbating its impact on YOLOv7 detection performance.

\subsubsection{Possible Defenses: Input Transformations}
We investigate simple defense strategies for mitigating adversarial asphalt art attacks by suppressing either color cues or texture cues, which may reduce the patch’s effectiveness. We evaluate three lightweight input transformations: grayscale, desaturation, and Gaussian blur. For grayscale, we convert the input image to grayscale and replicate the single channel to three channels for the detector. For desaturation, we reduce the image saturation by scaling the saturation channel to 0.5. For Gaussian blur, we apply a Gaussian filter with radius 5 to the input image before detection. Similarly, we apply these input transformation strategies to four asphalt art scenarios for comparison: clean, Color 1, Art 1, and Fishes.

\begin{table}[]
    \centering
    \caption{~Comparison of different input transformation defense strategies.}
    \label{tab:simple_defense}
    \begin{tabular}{c|c|c|c|c|c|c|c}
    \hline
        Arts & Defenses & max-F1 & Precision & Recall & mAP & mAP@0.5 & FDR \\
        \hline
        \multirow{4}{*}{Clean} & No defense & 0.995 & 0.991 & 1.0 & 0.723 & 0.999 & 0.009 \\
        \cline{2-8} 
         & Grayscale & 0.995 & 0.991 & 1.0 & 0.712 & 0.998 & 0.009 \\
         & Desaturation & 0.995 & 0.991 & 1.0 & 0.710 & 0.995 & 0.009 \\
         & Gaussian blur & 0.927 & 0.989 & 0.873 & 0.571 & 0.972 & 0.010 \\
        \hline
        \multirow{4}{*}{Color 1} & No defense & 0.995 & 0.991 & 1.0 & 0.701 & 0.998 & 0.009 \\
        \cline{2-8} 
         & Grayscale & 0.991 & 0.982 & 1.0 & 0.709 & 0.998 & 0.009 \\
         & Desaturation & 0.995 & 0.991 & 1.0 & 0.705 & 0.998 & 0.009 \\
         & Gaussian blur & 0.852 & 0.867 & 0.836 & 0.521 & 0.889 & 0.000 \\
        \hline
        \multirow{4}{*}{Art 1} & No defense & 0.804 & 0.927 & 0.709 & 0.416 & 0.744 & 0.030 \\
        \cline{2-8} 
         & Grayscale & 0.831 & 0.886 & 0.782 & 0.454 & 0.808 & 0.085 \\
         & Desaturation & 0.874 & 0.977 & 0.791 & 0.503 & 0.859 & 0.012 \\
         & Gaussian blur & 0.276 & 0.498 & 0.191 & 0.063 & 0.159 & 0.111 \\
        \hline
        \multirow{4}{*}{Fishes} & No defense & 0.679 & 0.983 & 0.518 & 0.373 & 0.641 & 0.544 \\
        \cline{2-8} 
         & Grayscale & 0.626 & 0.961 & 0.464 & 0.328 & 0.581 & 0.561 \\
         & Desaturation & 0.717 & 0.984 & 0.564 & 0.409 & 0.703 & 0.528 \\
         & Gaussian blur & 0.347 & 0.591 & 0.245 & 0.091 & 0.216 & 0.000 \\
        \hline
    \end{tabular}
\end{table}

The results are summarized in Table~\ref{tab:simple_defense}. On clean images and the mildest benign asphalt art Color 1, grayscale and desaturation have little impact on YOLOv7 detection performance, while Gaussian blur noticeably degrades performance (\eg, clean mAP from 0.723 to 0.571, Color 1 mAP from 0.701 to 0.521). For the most disruptive benign asphalt art Art 1, grayscale and desaturation provide partial mitigation (\eg, grayscale restores mAP from 0.416 to 0.454, and desaturation restores it to 0.503), whereas Gaussian blur collapses performance (mAP 0.063). For the malicious art Fishes, grayscale does not help improve pedestrian detection performance, desaturation yields modest improvement (mAP from 0.373 to 0.409), and Gaussian blur remains harmful to the final detection performance. Overall, desaturation is the most consistent lightweight transformation, but it still cannot restore performance to the clean level under strong or malicious asphalt art. These results show that adversarial asphalt art attacks remain effective under common color/texture preprocessing, indicating that their impact is not solely dependent on color or fine-grained texture. This finding suggests that more advanced defense strategies beyond simple image preprocessing are required to improve robustness in real-world scenarios.

\subsubsection{Possible Defenses: Adversarial Training}
Another effective defense strategy against adversarial attacks is adversarial training. We consider a realistic black-box threat model with unknown adversarial patterns, where adversaries deploy specially designed street art to degrade pedestrian detection, while a cyber threat defender for this application does not know the exact attack but has access to benign asphalt art patterns. Under this setting, cyber threat defenders can augment the training data with existing benign patterns to improve robustness against unseen malicious designs. Specifically, we incorporate simulated images with the most disruptive benign asphalt art, Art 1, together with clean data for training. The trained model is then evaluated on the malicious asphalt art pattern Fishes.

Table~\ref{tab:adv_training} shows the effectiveness of adversarial training. Without defense, the model suffers significant performance degradation, with a low recall and mAP. With adversarial training, detection performance improves substantially across multiple metrics. In particular, the max-F1 score rises from 0.679 to 0.963, and recall increases from 0.518 to 0.954. However, despite these improvements, the mAP metric score for our pedestrian detection model after adversarial training remains notably lower than that of the baseline model (\ie, 0.550 after adversarial training versus 0.723 for the baseline), suggesting that the adversarial defense does not fully restore performance. Additionally, the FDR remains high as 0.575, indicating that false detections are still prevalent. Overall, these results show that while adversarial training with benign asphalt art patterns can partially mitigate the attack, it is insufficient to completely eliminate its impact.

\begin{table}[]
    \centering
    \caption{~Effectiveness of adversarial training in defending malicious asphalt art Fishes.}
    \label{tab:adv_training}
    \begin{tabular}{c|c|c|c|c|c|c|c}
    \hline
        Arts & Defenses & max-F1 & Precision & Recall & mAP & mAP@0.5 & FDR \\
        \hline
        \multirow{2}{*}{Fishes} & No defense & 0.679 & 0.983 & 0.518 & 0.373 & 0.641 & 0.544 \\
        \cline{2-8} 
         & Adv-Training & 0.963 & 0.972 & 0.954 & 0.550 & 0.973 & 0.575 \\
        \hline
    \end{tabular}
\end{table}

\section{Discussion}
Our study reveals that while asphalt art offers aesthetic and community benefits, it can also adversely affect vision-based pedestrian detection systems. Our experiments demonstrate that certain designs, whether accidental or deliberately crafted, can degrade YOLOv7’s performance, highlighting a potential safety concern for roadway surveillance systems operating in visually complex urban environments.

\subsection{Limitations}
However, this work has several limitations. Firstly, all simulations were performed in a digital environment. This design choice was primarily motivated by the lack of publicly available datasets that capture road scenes containing asphalt art or murals alongside pedestrian activity under controlled, labeled conditions. To the best of our knowledge, no existing benchmark dataset includes such scenarios with sufficient diversity in art styles, placements, and environmental variations required for systematic analysis. Furthermore, collecting real-world data for this purpose is non-trivial and potentially unsafe, as it would require physically painting or modifying road surfaces. Such interventions may introduce unintended risks to existing surveillance systems before their impact is fully understood. Therefore, directly deploying untested asphalt art in real environments solely for data collection and/or field evaluation raises safety concerns. In this context, synthetic data provides a controlled, risk-free alternative, enabling us to systematically study the interaction between asphalt art and pedestrian detection models without altering real-world infrastructure. It enables precise manipulation of variables such as art style, placement, and density, which would be difficult to achieve consistently in real-world settings.

Despite these advantages, synthetic images may not fully capture real-world complexities. As a result, the observed performance changes may not fully generalize to physical deployments. However, the synthetic framework serves as a necessary first step toward understanding potential risks and guiding safer real-world experimentation. Based on these findings, future work can proceed toward controlled field deployments or curated real-world data collection, such as temporary installations in testbed environments, to empirically assess the real-world impact of various asphalt art styles on pedestrian detection performance. Additionally, our adversarial art creation relied on a heuristic design process guided by visual insights rather than a formal optimization framework, which may not represent the most effective or efficient attack strategies. Future work should focus on developing principled optimization methods for adversarial art design, validating the findings in real-world settings, and evaluating object detection robustness across a broader range of environmental and operational conditions.

\subsection{Implications}
Importantly, these findings raise broader implications for transportation agencies and urban planners. As cities increasingly embrace creative street designs to enhance visual appeal, community identity, and pedestrian engagement, it becomes crucial to consider the unintended consequences such designs may pose for vision perception systems that support public safety, such as those used in autonomous vehicles and smart surveillance infrastructure. As these AI-driven systems become more prevalent in urban environments, ensuring that street-level visual elements are interpretable not only to humans but also to machines will be essential for building safe, secure, and technologically compatible public spaces.

One takeaway from our study is the need to develop clear design guidelines for "AI-safe" asphalt art. These guidelines should aim to preserve the artistic and cultural value of street murals while minimizing the risk of interference with automated pedestrian detection systems. For example, certain color schemes, shapes, or contrast levels may be particularly disruptive to vision models and could be avoided or modified in the design phase. Collaborations between artists, engineers, and city planners could help create asphalt art that is both visually engaging and functionally safe, aligning with broader smart city goals. Establishing such interdisciplinary standards will be critical as urban spaces continue to evolve alongside emerging technologies.

\subsection{Countermeasures}
In our ablation study, we showed that input transformation techniques provide only limited effectiveness in mitigating the impact of asphalt art. While adversarial training improves YOLOv7 detection performance, it remains insufficient to fully eliminate the attack impact, particularly in reducing erroneous detections, as reflected by the persistently high FDR. These results highlight the need for more advanced and comprehensive defense strategies against both benign and malicious asphalt art patterns. One promising direction is multimodal perception, where camera inputs are complemented with LiDAR, radar, or depth sensing, thereby reducing the reliance on visual cues that can be easily manipulated. Additionally, incorporating temporal consistency or tracking across frames may further improve robustness against localized visual perturbations. Furthermore, periodic post-deployment updates that integrate data from newly introduced street art can help maintain long-term system reliability as urban environments evolve. Together, these directions provide promising avenues for future research in defending against asphalt art-induced vulnerabilities.

\section{Conclusions}
 Asphalt art, particularly artistic designs on pedestrian crosswalks, has gained significant popularity for enhancing crosswalk visibility and improving pedestrian safety. However, their impact on vision-based pedestrian surveillance systems remains largely unexplored. In this work, we demonstrate how these visually appealing street artworks can affect the performance of state-of-the-art pedestrian detectors. We investigate the effects of several real-world, benign asphalt art designs on pedestrian crosswalks using a series of controlled, digital-world simulations on a real-world pedestrian surveillance testbed. Our findings show that simple color variations have a negligible effect, whereas complex patterns can substantially degrade overall detection performance. In our experiments, the most impactful benign design reduced recall by 29.1\% and mAP@0.5 by 25.5\% compared to baseline detection results.
 
 We further examine how adversarially generated, malicious digital asphalt artworks on pedestrian crosswalks can impair pedestrian surveillance systems. Using a heuristic, ChatGPT-driven art generation loop, we crafted adversarial patterns that not only obscure real pedestrians but also induce false positive detections, reducing recall by 48.2\%, lowering mAP@0.5 by 35.8\%, and increasing the false discovery rate from 0.9\% to 54.4\%. By revealing how both aesthetic crosswalk designs and targeted adversarial patterns can undermine modern surveillance-based perception methods, this study highlights the need for more resilient designs of roadway signs and markings for the next-generation intelligent transportation systems surveillance systems.
\section{Acknowledgements}
This work is based upon the work supported by the National Center for Transportation Cybersecurity and Resiliency (TraCR) (a US Department of Transportation National University Transportation Center) headquartered at Clemson University, Clemson, South Carolina, USA. Any opinions, findings, conclusions, and recommendations expressed in this material are those of the author(s) and do not necessarily reflect the views of TraCR, and the US Government assumes no liability for the contents or use thereof.

In our experiments, we leveraged the ChatGPT image tool exclusively to assist in extracting real‐world street art patterns from photographs of asphalt installations, and to guide the design of our malicious adversarial crosswalk art. For manuscript preparation, ChatGPT was used only to help improve grammar. No information, text, figure, or table has been generated, nor has any kind of analysis been conducted using any Large Language Model or Generative Artificial Intelligence.

\section{Author Contributions}

The authors confirm their contribution to the paper: J. Ma, A. Enan, L. Cheng, M. Chowdhury. All authors reviewed the results and approved the final version of the manuscript. J. Ma and A. Enan are co-first authors; they have contributed equally.
\section{Conflict of Interest}
The authors declare no potential conflicts of interest with respect to the research, authorship, and/or publication of this article.

\newpage

\bibliographystyle{trb}
\bibliography{trb_template}
\end{document}